\documentclass[runningheads]{llncs}

\usepackage{eccv}

\usepackage{eccvabbrv}

\usepackage{graphicx}
\usepackage{booktabs}
\usepackage[utf8]{inputenc}
\usepackage{multirow}
\usepackage[table]{xcolor}
\definecolor{mygray}{gray}{0.9}
\definecolor{tabgray}{gray}{0.9}
\usepackage[accsupp]{axessibility}  %
\usepackage{wrapfig}
\usepackage{tabularx}
\usepackage{array}
\usepackage{pifont}
\newcolumntype{Y}{>{\centering\arraybackslash}X}

\usepackage{hyperref}

\usepackage{orcidlink}

\makeatletter
\renewcommand*{\@fnsymbol}[1]{\ensuremath{\ifcase#1\or *\or \dagger\or \ddagger\or
    \mathsection\or \mathparagraph\or \|\or **\or \dagger\dagger
    \or \ddagger\ddagger \else\@ctrerr\fi}}
\makeatother

\begin{document}

\title{PS-MOT: Cultivating Instance Awareness from Point Seeds for Multi-Object Tracking} 

\titlerunning{PS-MOT}

\author{Kai Luo\inst{1}\orcidlink{0000-0002-6433-905X} \and
Fei Teng\inst{1}\orcidlink{0009-0009-0894-3998} \and
Mengfei Duan\inst{1}\orcidlink{0000-0003-3017-9753} \and
Wanjun Jia\inst{1}\orcidlink{0009-0001-9541-8869} \and
Xu Wang\inst{1}\orcidlink{0009-0005-7982-0726} \and
Hao Shi\inst{2,3}\orcidlink{0000-0003-0184-2245} \and
Kunyu Peng\inst{4,5}\orcidlink{0000-0002-5419-9292} \and
Zhiyong Li\inst{1}\orcidlink{0000-0001-9720-5915} \and
Kailun Yang\inst{1,}\thanks{Correspondence: kailun.yang@hnu.edu.cn}\orcidlink{0000-0002-1090-667X}
}

\authorrunning{K.~Luo~\textit{et al.}}

\institute{
Hunan University, China \and
Ant Group, China \and
Zhejiang University, China \and
Karlsruhe Institute of Technology, Germany \and
INSAIT, Sofia University ``St. Kliment Ohridski'', Bulgaria
}

\maketitle

\begin{abstract}
We introduce Point-supervised Multi-Object Tracking (PS-MOT) as a cost-effective alternative to traditional bounding box supervision, shifting the focus from spatial fitting to topological center-driven representation. However, PS-MOT faces challenges, \eg, spatial ambiguity and identity drift due to the lack of explicit geometric structure and scale constraints. To address these, we propose PS-Track, a hierarchical pipeline transitioning from \textit{points} to \textit{instances} across data, model, and loss levels. At the data level, we introduce Temporal-Feedback Prompting (TFP) to evolve points into temporally consistent pseudo-labels using negative spatial cues and motion priors. At the model level, we design the Point-Excited Wavelet Attention (PEWA) module, which leverages semantic correlations to activate high-frequency components, ``hallucinating'' object boundaries. At the loss level, Uncertainty-Guided Gaussian Learning (UGL) models pseudo-labels as probabilistic distributions, dynamically calibrating supervision intensity. Experiments on DanceTrack, EmboTrack, SportsMOT, and JRDB demonstrate that PS-Track provides a feasible and effective point-supervised alternative across diverse tracking scenarios, establishing a new state-of-the-art for point-supervised tracking. The source code is available at \url{https://github.com/xifen523/PS-MOT}.

\end{abstract}

\section{Introduction}
\label{sec:intro}

Multi-Object Tracking (MOT) has long served as a cornerstone of spatio-temporal perception, empowering autonomous systems to maintain persistent identity consistency and trajectory continuity for multiple targets in dynamic environments~\cite{hu2023planning, zhu2025sparsead}. Despite the rapid evolution of Tracking-By-Detection (TBD) paradigms~\cite{yi2024ucmctrack,shim2025focusing} and query-based~\cite{luo2025omnitrack++, gao2023memotr} Transformer architectures, which have significantly bolstered tracking precision, contemporary research remains constrained by a formidable scalability bottleneck.

\begin{figure}[!t]
  \centering
  \includegraphics[width=\textwidth]{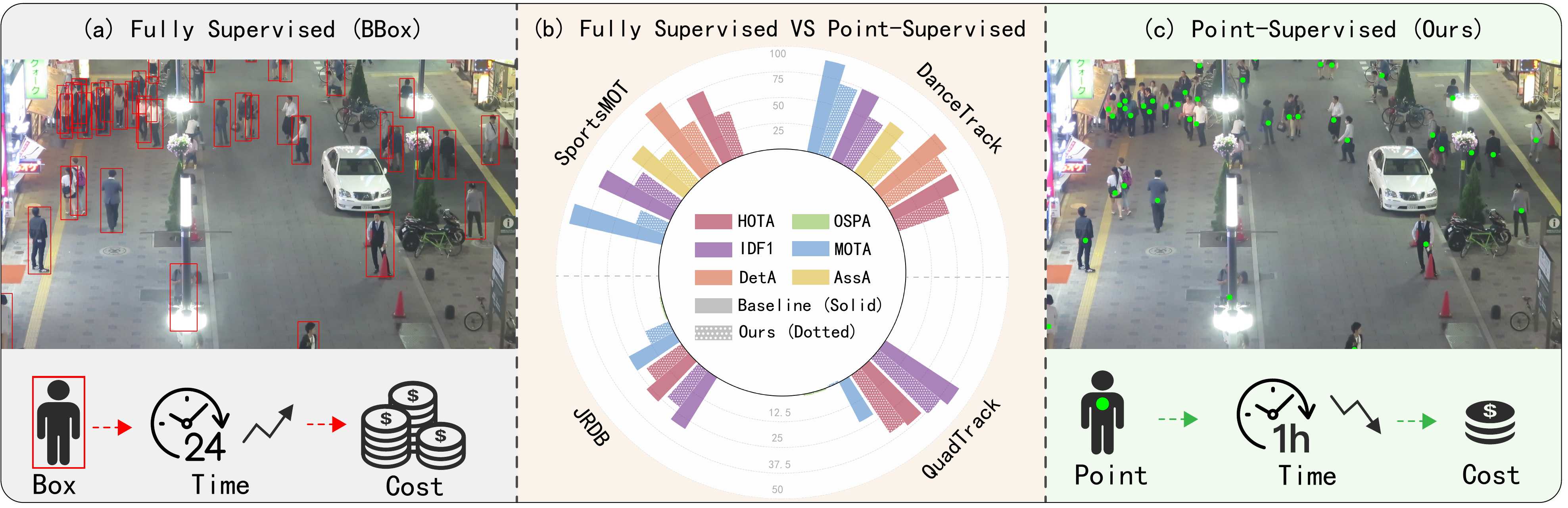}
  \vskip -2ex
  \caption{\textbf{Comparison between Fully Supervised MOT and our proposed Point-Supervised MOT (PS-MOT)}. 
  (a) Traditional MOT requires labor-intensive bounding box annotations, resulting in high time and labor costs.
(b) Performance comparison between fully supervised and point-supervised methods (solid bars: fully supervised baseline; dotted bars: ours). (c) Our point-supervised paradigm adopts a topological center-driven representation via simple point prompts (green dots), significantly reducing annotation time and labor (\textit{e.g.}, 0.7-0.9s per point \textit{vs.} 7-10s per box)~\cite{cheng2022pointly, bearman2016s, papadopoulos2017extreme}.}
\label{fig:teaser}
\vskip-4ex
\end{figure}

State-of-the-art MOT methods~\cite{Gao_2025_CVPR, lv2024diffmot} rely heavily on large-scale, frame-by-frame dense bounding box annotations, as illustrated in Fig.~\ref{fig:teaser} (a). 
This dependency presents a dual challenge. 
First, bounding box annotation necessitates precise alignment between object boundaries and camera perspectives. In complex scenes, rigid axis-aligned boxes often struggle to accurately depict true geometric structures, rendering precise labeling both arduous and ambiguous. Second, establishing temporal associations across frames requires maintaining identity consistency and trajectory continuity, which further exacerbates the manual annotation burden~\cite{zhou2024openannotate2,dendorfer2020mot20}. 
Consequently, constructing millions of bounding boxes is prohibitively expensive, confining MOT research to a few meticulously curated data domains and hindering its scalability in real-world applications~\cite{mao2021one}. 
To break this barrier, we propose a minimalist supervision paradigm: Point-supervised Multi-Object Tracking (PS-MOT), as illustrated in Fig.~\ref{fig:teaser} (c).

However, adopting point supervision introduces a fundamental precision-ambiguity paradox. A scale-less point naturally lacks geometric structure and scale constraints, rendering the supervision signal inherently ambiguous and underdetermined in the spatial domain~\cite{zhang2025point2rbox,yu2025point2rbox}. This uncertainty triggers two critical issues in temporal modeling: first, spatial leakage occurs as the model, lacking explicit boundary constraints and a physical understanding of spatial occupancy, allows supervision signals to permeate into the background or adjacent targets in crowded scenes, thereby weakening spatial discriminative power; second, identity drift arises because the uncertainty in scale representation forces the model to implicitly infer target extents, leading to unstable cross-frame associations, temporal jitter, and frequent ID-switches.

To address these challenges, we propose PS-Track, a unified framework that orchestrates a coarse-to-fine instance evolution across the data, model, and loss levels. Our core philosophy hinges on the insight that while point annotations lack explicit spatial cognition, the synergy between foundation model priors and temporal dynamics can effectively recover missing scale representations within a probabilistic space. This enables a progressive transition from a mere topological center to a precise and identity-consistent instance representation, effectively bridging the gap between sparse supervision and dense perception requirements.

This evolutionary process is operationalized through three synergistic components. At the data level, Temporal-Feedback Prompting (TFP) evolves static point seeds into temporally consistent pseudo-labels by injecting negative spatial cues and motion priors, thereby resolving identity merging and fragmentation in crowded scenes.
Parallel to this data evolution, the Point-Excited Wavelet Attention (PEWA) module at the model level directly treats the original annotated points as frequency exciters. Inspired by the top-down mechanism of the human visual system~\cite{bar2003cortical,ullman1995sequence}, PEWA leverages these precise point seeds to selectively activate high-frequency wavelet coefficients, effectively hallucinating sharp object boundaries from sparse signals. Finally, to mitigate the intrinsic noise of evolved 
\begin{wrapfigure}{r}{0.3\textwidth}
    \vskip -4ex
    \centering
    \includegraphics[trim=0mm 0mm 0mm 0mm, clip, width=\linewidth]{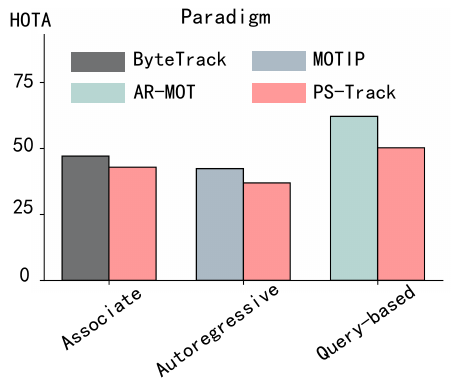}
    \vskip -2ex
    \caption{Versatility of PS-Track across mainstream tracking paradigms.}
    \label{fig:paradigm}
    \vskip -5ex
\end{wrapfigure}
labels, we transition from deterministic regression to Uncertainty-Guided Gaussian Learning (UGL) at the loss level. Since point annotations lack physical extents, deterministic box regression causes noise memorization. 
UGL circumvents this by modeling pseudo-labels as Gaussian distributions, naturally accommodating spatial ambiguity. 
To mitigate label noise, UGL dynamically down-weights unreliable targets using the joint quality score generated by TFP. This data-to-loss synergy ensures precise and robust convergence.

As shown in Fig.~\ref{fig:teaser}(b), PS-Track demonstrates promising performance under point supervision, narrowing the gap to fully supervised state-of-the-art models~\cite{Gao_2025_CVPR,luo2025omnitrack++} in certain scenarios while generalizing from standard pinhole cameras~\cite{sun2022dancetrack,cui2023sportsmot} to embodied panoramic vision~\cite{martin2021jrdb,luo2025omnitrack++}. Furthermore, Fig.~\ref{fig:paradigm} demonstrates its architectural versatility. Our framework can be integrated into diverse tracking paradigms—including TBD, autoregressive, and query-based models—maintaining competitive accuracy (\textit{e.g.}, trailing fully supervised~\cite{zhang2022bytetrack} by merely $4.2$ HOTA on DanceTrack~\cite{sun2022dancetrack} val set). 
Ultimately, this cross-scenario and cross-paradigm adaptability establishes PS-Track as a universal foundation for the novel PS-MOT~task.

Our contributions are summarized as follows:
\begin{itemize}
\item \textbf{Formalization of the PS-MOT Paradigm:}  
We formulate the task of Point-supervised MOT (PS-MOT), offering a scalable alternative to traditional para\-digms by shifting the focus from labor-intensive spatial fitting to a topological center-driven representation. 
This perspective effectively bypasses the scalability bottleneck inherent in bounding box-based methods.
\item \textbf{A Hierarchical Evolutionary Framework:} 
We develop PS-Track, a unified pipeline that orchestrates a coarse-to-fine \textit{point-to-instance} transition by synergistically integrating temporal label evolution (TFP), frequency-domain boundary hallucination (PEWA), and uncertainty-aware probabilistic learning. This hierarchical design enables stable instance representation to emerge from sparse, scale-less point seeds.
\item \textbf{Empirical Validation under Sparse Supervision:}
Extensive experiments on four challenging benchmarks—DanceTrack~\cite{sun2022dancetrack}, EmboTrack~\cite{luo2025omnitrack++}, SportsMOT~\cite{cui2023sportsmot}, and JRDB~\cite{martin2021jrdb}—demonstrate that PS-Track achieves performance competitive with fully supervised counterparts. Remarkably, our method establishes a new state-of-the-art for point-supervised tracking while reducing annotation time by roughly $64\%$ compared to full supervision.
\end{itemize}

\section{Related Work}
\noindent\textbf{Annotation Paradigms in MOT.} Current MOT is dominated by fully supervised paradigms, including Tracking-by-Detection (TBD)~\cite{yi2024ucmctrack, yang2024hybrid, shim2025focusing} with precise detectors~\cite{yolox2021,peng2024dfine}, and end-to-end transformers like MOTR~\cite{zeng2021motr}, TrackFormer~\cite{meinhardt2022trackformer}, and MOTIP~\cite{Gao_2025_CVPR}. Despite state-of-the-art performance on standard benchmarks~\cite{dendorfer2021motchallenge, dendorfer2020mot20, sun2022dancetrack, cui2023sportsmot}, their success heavily relies on massive bounding box ($BBox$) annotations. This dependence creates a severe scalability barrier in embodied and panoramic scenarios~\cite{martin2021jrdb, cui2023sportsmot}. In such densely crowded and geometrically distorted views, annotating precise $BBox$es becomes prohibitively labor-intensive and expensive, effectively bottlenecking the curation of large-scale datasets. Conversely, point-level signals offer a highly efficient alternative, providing stable semantic anchors that drastically reduce the annotation burden while naturally resisting perspective distortions. To mitigate annotation costs, various label-efficient tracking strategies have emerged. However, unsupervised~\cite{he2021learnable, li2025self} and semi-supervised~\cite{zhao2025crtrack, wang20243d} methods often struggle with occlusions or rely heavily on initial seed quality, while weakly-supervised approaches~\cite{lu2024self, lim2024track} still depend on $BBox$ priors for scale estimation. Although point supervision excels in static object detection~\cite{P2BNet, CPR}, its application in MOT remains under-explored. PS-Track addresses this by shifting the focus from absolute spatial precision to temporal identity consistency, offering a highly scalable and robust tracking solution without requiring any dense bounding box annotations.

\noindent\textbf{Weakly-Supervised Localization.}
Weakly-Supervised Localization (WSL) serves as a critical bridge between coarse supervisory signals, \eg, image-level tags~\cite{Bilen16, tang2017multiple}, scribbles~\cite{gao2025scribble, hayat2025superpixel}, and sparse points~\cite{zheng2025seg2box, chan2025sparse}, and dense instance-level understanding. 
Within the point-supervised regime, research has historically focused on static object detection, where frameworks like P2BNet~\cite{P2BNet} and its subsequent iterations~\cite{yu2024point2rbox, yu2025point2rbox} utilize Multiple Instance Learning (MIL) or coarse-to-fine regression to hallucinate bounding boxes from dimensionless point seeds. While these methods demonstrate impressive localization accuracy in static images, they inherently overlook the temporal continuity essential for video sequences. In the context of MOT~\cite{yi2024ucmctrack, luo2025omnitrack++}, independent frame-wise scale estimation frequently induces temporal jitter and identity instability. Consequently, reformulating point-to-mask generation as a trajectory-aware process, guided by motion priors to enforce spatio-temporal coherence, emerges as a necessary evolution to ensure tracking stability under sparse supervision.

A key challenge in MOT is the Location-Boundary Mismatch, where point seeds provide precise semantic centering but lack geometric scale or edge information. Conventional convolutional architectures~\cite{he2016deep, he2017mask}, limited by local receptive fields and spatial pooling, struggle to reconstruct sharp boundaries from sparse points. Recent frequency-domain learning~\cite{tan2024frequency, zheng2022learning} shows that high-frequency wavelet coefficients better represent structural edges than spatial intensities. By using annotated points to activate these high-frequency components—drawing inspiration from wavelet-based feature decomposition~\cite{shu2026waveformer}—precise boundaries can be hallucinated without excessive architectural complexity. Additionally, weak supervision introduces heteroscedastic noise in pseudo-labels, and treating them as deterministic ground truth can hinder convergence. Shifting to probabilistic distribution learning, based on aleatoric uncertainty modeling~\cite{kendall2017uncertainties, gal2016dropout}, allows the framework to adaptively down-weight unreliable signals, which stabilizes gradients from inaccurate labels to ensure reliable model convergence.

\begin{figure}[!t]
  \centering
  \includegraphics[trim=0mm 0mm 0mm 0mm, clip, width=\linewidth]{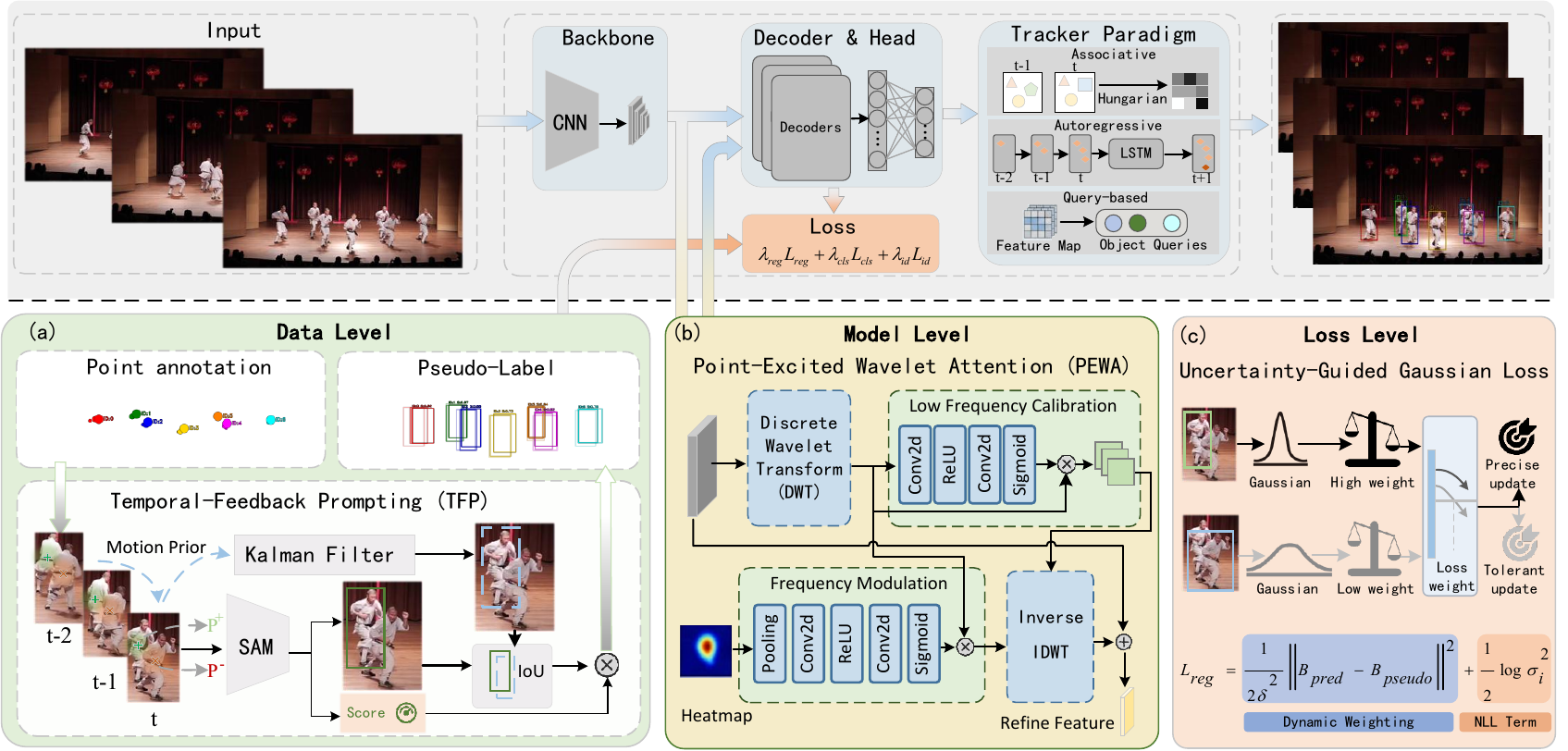} %
  \vskip -2ex
  \caption{The overview of our proposed framework. It operates on a coarse-to-fine paradigm across three levels: (a) \textbf{Data Level}: The Temporal-Feedback Prompting mechanism evolves sparse points into consistent pseudo-labels; (b) \textbf{Model Level}: The Point-Excited Wavelet Attention (PEWA) module leverages frequency decomposition to Hallucinate boundaries from point cues; (c) \textbf{Loss Level}: The Uncertainty-Guided Gaussian Loss dynamically re-weights supervision signals based on temporal reliability.}
  \label{fig:overall}
  \vskip -4ex
\end{figure}

\section{Methodology}

\subsection{Overall Architecture}

Point-supervised Multi-Object Tracking (MOT) presents a fundamental paradox: the supervision signal (a single pixel coordinate) is deterministic in location but highly ambiguous in scale and boundary. Traditional methods treat these points as static anchors, leading to semantic drift during occlusion and identity switching in crowded scenes. To bridge the gap between sparse supervision and dense pixel-level understanding, we propose a unified framework that treats the annotated point not merely as a label, but as a \textit{dynamic exciter} of semantic regions, enabling progressive scale inference and boundary reconstruction.
Our framework systematically addresses the ambiguity of point supervision through a three-stage Evolution-Perception-Adaptation paradigm, as illustrated in Fig.~\ref{fig:overall}, spanning data, model, and loss levels:

\begin{itemize}
    \item \textbf{Data Level: Temporal-Feedback Prompting (Evolution).} 
    We argue that a single frame is insufficient to resolve the ambiguity of a point. 
    By introducing a closed-loop interaction with the Segment Anything Model (SAM)~\cite{carion2025sam3segmentconcepts}, we transform static point annotations into temporally consistent pseudo-bounding boxes. Unlike naive generation, our module incorporates negative spatial cues to separate closely interacting targets and leverages motion-guided box constraints to suppress segmentation fragmentation, effectively evolving noisy points into reliable trajectory priors.

    \item \textbf{Model Level: Point-Excited Wavelet Attention (Perception).} 
    How can a network hallucinate accurate boundaries from a single point? Inspired by the top-down mechanism of the human visual system, where coarse global information guides local detail processing~\cite{bar2003cortical,ullman1995sequence}, we introduce the Point-Excited Wavelet Attention (PEWA) module. Leveraging Discrete Wavelet Transform (DWT)~\cite{shensa2002discrete}, PEWA decomposes features into low-frequency (global context) and high-frequency (local boundary) components. The annotated point acts as a frequency exciter, selectively activating high-frequency coefficients in the wavelet domain to reconstruct sharp object boundaries without heavy upsampling parameters~\cite{shu2026waveformer}.

    \item \textbf{Loss Level: Uncertainty-Guided Gaussian Learning (Adaptation).} 
    Acknowledging that pseudo-labels inherently contain noise, we shift from deterministic regression to probabilistic modeling. We interpret the pseudo-labels as observations with heteroscedastic noise and formulate an uncertainty-aware loss function based on a Gaussian likelihood. 
    By inheriting the Joint Quality Score from the data level—which fuses visual confidence and temporal consistency—the network dynamically down-weights unreliable samples, achieving robust convergence despite label noise.
\end{itemize}

\begin{figure}[!t]
  \centering
  \includegraphics[trim=0mm 0mm 2mm 0mm, clip, width=\textwidth]{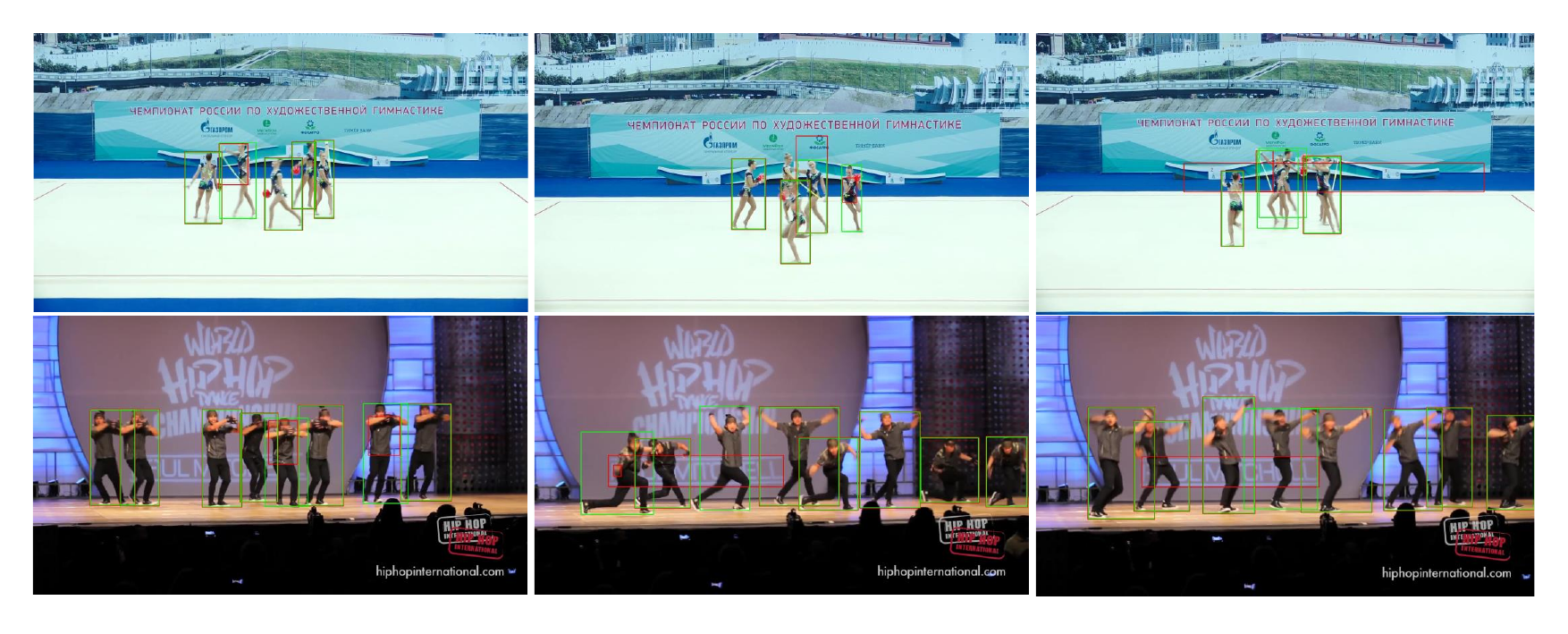}
  \vskip -4ex
  \caption{\textbf{Visual comparison of pseudo-labels generated by the original SAM and our Temporal-Feedback Prompting (TFP) module.} 
  The \textcolor{red}{red box} represents the initial pseudo-labels from SAM~\cite{carion2025sam3segmentconcepts}, whereas the \textcolor{green}{green box}  denotes the labels refined by TFP. As illustrated, the TFP-generated labels exhibit significantly higher accuracy and robustness, effectively correcting the misalignments in the original SAM output.}
  \label{fig:visual_comparison_pseudo_labels}
  \vskip -4ex
\end{figure}

\subsection{Data Level: Temporal-Feedback Prompting (TFP)}
\label{sec:data_level}

The efficacy of point-supervised learning is heavily contingent on the quality of the generated pseudo-labels. A naive application of foundation models (\textit{e.g.}, SAM) often fails in dynamic tracking scenarios due to two critical issues: \textit{Identity Merging}, where a single mask erroneously encompasses multiple crowded targets, and \textit{Semantic Fragmentation}, where only a discriminative part (\textit{e.g.}, a limb) is segmented due to occlusion as shown in Fig. \ref{fig:visual_comparison_pseudo_labels}. To mitigate these, we propose the Temporal-Feedback Prompting (TFP) mechanism, which evolves sparse points into reliable bounding boxes through spatio-temporal constraints.

\noindent\textbf{Spatial Disambiguation via Negative Cues.} 
In crowded scenes like DanceTrack, targets often share overlapping boundaries, confusing the segmentation model when prompted with a single positive point $p^+_i$. To enforce instance separation, we introduce an \textit{Interactive Negative Cues} strategy. For a target $i$, we construct a set of negative prompts $\mathcal{P}^-_i$ by sampling the centroids of its spatial neighbors:
\begin{equation}
    \mathcal{P}^-_i = \{ p^+_j \mid j \in \mathcal{N}(i), \; \| p^+_i - p^+_j \|_2 < \tau_{dist} \},
\end{equation}
where $\mathcal{N}(i)$ denotes the set of co-existing tracks in the current frame and $\tau_{dist}$ is a proximity threshold. By explicitly feeding $\{p^+_i, \mathcal{P}^-_i\}$ into SAM, we impose a semantic firewall, compelling the model to reject features belonging to adjacent identities, thereby resolving the identity merging dilemma.

\noindent\textbf{Temporal Regularization with Motion Priors.} 
To combat segmentation fragmentation caused by occlusion, we leverage the temporal continuity of object trajectories. 
We maintain a motion state estimate for each tracklet using a Kalman filter. 
Before segmentation, we project the tracklet's state to the current frame $t$ to obtain a motion prior box $B_{t|t-1}$. 
This prior serves as a spatial constraint for the foundation model:
\begin{equation}
    M_t = \text{SAM}(I_t, \text{point}=\{p^+_i, \mathcal{P}^-_i\}, \text{box}=B_{t|t-1}),
\end{equation}
where the inclusion of $B_{t|t-1}$ restricts the segmentation search space, effectively suppressing outlier regions and ensuring the scale consistency of the generated mask $M_t$, 
thereby enhancing temporal stability and reducing drift.

\noindent\textbf{Joint Quality Scoring for Robust Learning.} 
Unlike traditional methods that treat all pseudo-labels as ground truth (hard labeling), we advocate for a soft labeling approach. We compute a Joint Quality Score $S_{joint}$ for each generated pseudo-box $B_{pseudo}$ to quantify its reliability:
\begin{equation}
    S_{joint} = S_{sam} \cdot (\alpha \cdot \text{IoU}(B_{pseudo}, B_{t|t-1}) + \beta),
\end{equation}
where $S_{sam}$ is the model's native visual confidence, and the IoU term measures the consistency between the visual segmentation and the physical motion prior. $\alpha$ and $\beta$ are scaling hyperparameters that control the contribution and base offset of temporal consistency, respectively. This score $S_{joint}$ is preserved and utilized in the subsequent loss level to dynamically down-weight unreliable samples, creating a noise-tolerant supervision signal for stable and robust probabilistic optimization dynamics.

\subsection{Model Level: Point-Excited Wavelet Attention (PEWA)}
\label{sec:model_level}
While the Data Level evolves points into pseudo-boxes, the network's internal representation must also adapt to the sparsity of point supervision. 
A fundamental challenge is the \textit{Location-Boundary Mismatch}: a point indicates where an object is, but the convolution kernels struggle to infer what shape it has without explicit boundary supervision. 
To bridge this gap, we draw inspiration from the Top-down processing mechanism of the human visual system~\cite{bar2003cortical}, where high-level abstract representations guide the perception of low-level details. 
We propose the Point-Excited Wavelet Attention (PEWA) module, which treats annotated points as frequency exciters during training to hallucinate object boundaries in the wavelet domain under sparse supervision. During inference, it is bypassed, preserving the standard MOT setting with only RGB frames as input.

\noindent\textbf{Wavelet Decomposition for Feature Decoupling.} 
Standard CNNs inherently entangle semantic and structural information, hindering boundary hallucination from sparse points. While WaveFormer~\cite{shu2026waveformer} explores frequency decoupling for general representation, we specifically exploit the Discrete Wavelet Transform (DWT) to isolate high-frequency signals for precise boundary reconstruction. By decoupling the feature map $X \in \mathbb{R}^{C \times H \times W}$ via a single-level Haar wavelet decomposition, we obtain:
\begin{equation}
    X_{LF}, \{X_{HF}^{k}\}_{k=1}^3 = \text{DWT}(X),
\end{equation}
where $X_{LF}$ represents the Low-Frequency approximation containing global semantic context (corresponding to the abstract representation in top-down theory~\cite{bar2003cortical}), and $\{X_{HF}^{k}\}$ (consisting of LH, HL, HH sub-bands) captures High-Frequency details such as edges and textures.

\noindent\textbf{Point-Guided Frequency Excitation.} 
In the decoupled domain, object boundaries are concentrated in the HF components. 
However, without supervision, these HF components are dominated by background clutter. 
We propose a \textit{Masked Frequency Modulation} strategy to suppress noise and amplify target-related edges.
First, we generate a spatial Gaussian heatmap $G \in \mathbb{R}^{1 \times H \times W}$ centered at the annotated point $p_i$, representing the potential object scope. This heatmap is downsampled to match the wavelet domain resolution. We then generate a \textit{Frequency Excitation Mask} $M_{exc}$ via a lightweight modulator $\phi$:
\begin{equation}
    M_{exc} = \sigma(\phi(G_{\downarrow})), \quad M_{exc} \in \mathbb{R}^{C \times \frac{H}{2} \times \frac{W}{2}},
\end{equation}
where $\sigma$ is the Sigmoid function. This mask acts as a biological spotlight, selectively activating the high-frequency coefficients around the target:
\begin{equation}
    \hat{X}_{HF}^k = X_{HF}^k \odot M_{exc}, \quad k \in \{LH, HL, HH\}.
\end{equation}
By this operation, we enforce a strong prior: only boundaries in the vicinity of the annotated point are valid within localized semantic regions.
The Low-Frequency component $X_{LF}$ is left unmodulated (or processed via standard self-attention) to preserve the global identity consistency.

\noindent\textbf{Inverse Reconstruction and Feature Fusion.} 
Finally, the modulated high-frequency details and the global context are recombined via the Inverse Discrete Wavelet Transform (IDWT) to reconstruct the refined feature $\hat{X}$:
\begin{equation}
    \hat{X} = \text{IDWT}(X_{LF}, \{\hat{X}_{HF}^k\}_{k=1}^3) + X.
\end{equation}
This reconstruction process is parameter-efficient compared to traditional upsampling layers. 
Through PEWA, the training process encourages boundary-sensitive feature responses from sparse point cues, which improves the discriminative embeddings used by the downstream tracker.

\subsection{Loss Level: Uncertainty-Guided Gaussian Learning (UGL)}
\label{sec:loss_level}
In the final stage of our framework, we address the inherent noise within the generated pseudo-labels. 
Standard MOT training objectives (\textit{e.g.}, $L_1$ or GIoU loss) implicitly assume that the ground truth follows a Dirac delta distribution (\textit{i.e.}, zero uncertainty). 
Applying such rigid constraints to our evolved pseudo-boxes would force the network to overfit to the noise introduced by SAM's segmentation errors or occlusion drifts. 
To fundamentally resolve this, we shift the supervision paradigm from deterministic regression to probabilistic likelihood maximization.

\noindent\textbf{Modeling Heteroscedastic Aleatoric Uncertainty.} 
We model each pseudo-bounding box $B_{pseudo}$ not as a fixed ground truth, but as an observation drawn from a Gaussian distribution with an underlying true mean $B_{gt}$ and a variance $\sigma^2$ representing the observation noise:
\begin{equation}
    B_{pseudo} \sim \mathcal{N}(B_{gt}, \sigma^2).
\end{equation}
Here, the variance $\sigma$ captures the \textit{aleatoric uncertainty} inherent in the weak supervision. 
Crucially, we propose that this uncertainty is inversely correlated with our Joint Quality Score ($S_{joint}$) derived in the Data Level (Sec.~\ref{sec:data_level}). 
Intuitively, a trajectory with high temporal consistency and visual confidence should have low variance (high precision). We thus map the score to the variance domain:
\begin{equation}
    \sigma_i = \frac{1}{S_{joint, i} + \epsilon},
\end{equation}
where $\epsilon$ is a stability constant.

\noindent\textbf{Gaussian Likelihood Loss (GLL).} 
Instead of minimizing the distance directly, we maximize the log-likelihood of the predicted box $B_{pred}$ given the noisy observation. 
This leads to our Uncertainty-Guided Regression Loss, formulated as the negative log-likelihood (NLL):
\begin{equation}
    \mathcal{L}_{reg} = \frac{1}{N} \sum_{i=1}^{N} \left( \frac{1}{2\sigma_i^2} \| B_{pred, i} - B_{pseudo, i} \|_2^2 + \frac{1}{2} \log \sigma_i^2 \right).
\end{equation}
This formulation offers a profound physical interpretation. The first term acts as a \textit{dynamic reweighting} mechanism: reliable samples (high $S_{joint}$, small $\sigma$) receive large weights ($\frac{1}{2\sigma^2}$) to enforce precise localization, whereas weights for unreliable samples (\textit{e.g.}, occluded targets) decay, allowing the model to tolerate potential labeling errors. 
Notably, since $\sigma_i$ is deterministically derived from the joint quality score $S_{joint,i}$, the second term serves as a probabilistic calibration term in the Gaussian NLL. It preserves the absolute likelihood scale across batches with varying pseudo-label quality, ensuring that the objective remains statistically meaningful rather than becoming a purely relative weighted regression loss.

\noindent\textbf{Total Objective.} 
The final training objective integrates the uncertainty-guided regression and soft classification components:
\begin{equation}
    \mathcal{L}_{total} = \lambda_{reg} \mathcal{L}_{reg} + \lambda_{cls} \mathcal{L}_{cls} + \lambda_{id} \mathcal{L}_{id},
\end{equation}
where $\mathcal{L}_{cls}$ is the standard classification loss, and $\mathcal{L}_{id}$ represents the identity prediction loss as defined in MOTIP~\cite{Gao_2025_CVPR}, supervised by generated identity tags.
Through the Uncertainty-Guided Gaussian Learning, PS-Track can adapt to the noise distribution of point-supervised data without manual label cleaning.

\begin{table}[h]
  \centering
  \vspace{-2em}
  \caption{Quantitative comparison on the DanceTrack~\cite{sun2022dancetrack} test set.}
  \vspace{-1em}
  \label{tab:dancetrack_val}
  \footnotesize 
  \setlength{\tabcolsep}{2pt} 

    \resizebox{.8\textwidth}{!}{
  \begin{tabular}{l|cc|ccccc}
    \toprule
    Methods & Venue & \hspace{1pt}Year \hspace{1pt}  & HOTA$\uparrow$ & DetA$\uparrow$ & AssA$\uparrow$ & MOTA$\uparrow$ & IDF1$\uparrow$ \\
    \midrule
    CenterTrack \cite{zhou2020tracking} & ECCV & 2020 & 41.8 & 78.1 & 22.6 & 86.8 & 35.7 \\
    FairMOT \cite{zhang2021fairmot}      & IJCV & 2021 & 39.7 & 66.7 & 23.8 & 82.2 & 40.8 \\
    QDTrack \cite{pang2021quasi}       & CVPR & 2021 & 54.2 & 80.1 & 36.8 & 87.7 & 50.4 \\
    TraDes \cite{wu2021track}        & CVPR & 2021 & 43.3 & 74.5 & 25.4 & 86.2 & 41.2 \\
    SUSHI \cite{cetintas2023unifying} & CVPR & 2023 & 63.3 & - & 50.1 & 88.7 & 63.4 \\
    FineTrack \cite{ren2023focus} & CVPR & 2023 & 52.7 & - & 38.5 & 89.9 & 59.8 \\
    OC-SORT \cite{cao2023observation} & CVPR & 2023 & 55.1 & 80.3 & 38.3 & 92.0 & 54.6 \\
    CMTrack \cite{shim2024confidence} & ICIP & 2024 & 61.8 & - & 46.4 & 92.5 & 63.3 \\
    UCMCTrack+ \cite{yi2024ucmctrack} & AAAI & 2024 & 63.6 & - & 51.3 & 88.9 & 65.0 \\
    TrackTrack \cite{shim2025focusing} & CVPR & 2025  & 66.5 & - & 52.9 & 93.6 & 67.8 \\
    MOTIP \cite{Gao_2025_CVPR} & CVPR & 2025  & 69.6 & 80.4 & 60.4 & 90.6 & 74.7 \\
    \rowcolor[gray]{0.95}
    PS-Track (ours) & - & - & 52.3 & 68.2 & 40.4 & 71.7 & 53.4 \\
    \bottomrule

  \end{tabular}
  }

  \vspace{-4em}
\end{table}
\section{Experiments}
\subsection{Setups}
\textbf{Datasets.} 
We evaluate PS-Track on four diverse benchmarks to verify its robustness against spatial ambiguity and semantic drift. 
DanceTrack~\cite{sun2022dancetrack} (105K frames) and SportsMOT~\cite{cui2023sportsmot} (150K frames) feature severe non-linear motions and uniform appearances, rigorously testing TFP's identity preservation. For embodied robotic scenarios, we utilize JRDB~\cite{martin2021jrdb} (panoramic dense crowds) and EmboTrack~\cite{luo2025omnitrack++} (extreme ego-motion from mobile platforms). 
Together, these datasets encompass complex locomotion dynamics and diverse noise distributions, comprehensively verifying PEWA's boundary hallucination and our UGL under strictly sparse point supervision. Detailed experimental settings are provided in the supplementary material.

\noindent\textbf{Metrics.} 
We evaluate tracking performance using HOTA~\cite{luiten2021hota}, along with DetA and AssA to assess boundary estimation and identity consistency. 
We also report MOTA~\cite{bernardin2008evaluating} and IDF1~\cite{ristani2016performance} for overall robustness and long-term stability. 
For embodied panoramic datasets (\textit{e.g.}, JRDB~\cite{martin2021jrdb}, EmboTrack~\cite{luo2025omnitrack++}), OPSA~\cite{martin2021jrdb} is additionally used to measure pattern-level consistency under sparse supervision.

\noindent\textbf{Implementation Details.}
PS-Track is built upon MOTIP~\cite{Gao_2025_CVPR} with Deformable DETR~\cite{zhu2020deformable} and an ImageNet-pretrained ResNet-50 backbone. All models are trained on a single NVIDIA GeForce RTX 5090 GPU with 30-frame video clips per batch. We optimize PS-Track with AdamW for $10$ epochs using a base learning rate of $1\times10^{-4}$, with the backbone learning rate scaled to $1\times10^{-5}$ and decayed at the 6-th and 9-th epochs. Compared with MOTIP, PS-Track introduces only moderate training overhead in our setting, increasing memory from $23.7$GB to $24.2$GB and training time from $7.4$h to $8.0$h per epoch, while keeping inference speed nearly unchanged since SAM and PEWA are bypassed at test time. During inference, PS-Track remains fully point-free and follows the standard MOT setting, requiring only RGB frames as input.

\begin{table*}[t] %
    \centering

    \begin{minipage}[t]{0.48\linewidth}
        \centering
        \caption{Quantitative comparison on the SportsMOT test set~\cite{cui2023sportsmot}.}
        \vspace{-1em}
        \label{tab:mot_benchmark}
        \footnotesize
        \setlength{\tabcolsep}{2pt}
        \resizebox{\linewidth}{!}{
            \begin{tabular}{l|ccccc}
                \toprule
                Methods & HOTA$\uparrow$ & DetA$\uparrow$ & AssA$\uparrow$ & MOTA$\uparrow$ & IDF1$\uparrow$ \\
                \midrule
                \multicolumn{6}{l}{\textit{w/o extra data:}} \\
                FairMOT~\cite{zhang2021fairmot} & 49.3 & 70.2 & 34.7 & 86.4 & 53.5\\
                QDTrack~\cite{pang2021quasi} & 60.4 & 77.5 & 47.2 & 90.1 & 62.3\\
                ByteTrack~\cite{zhang2022bytetrack} & 62.1 & 76.5 & 50.5 & 93.4 & 69.1 \\
                TrackFormer~\cite{meinhardt2022trackformer} & 63.3 & 66.0 & 61.1 & 74.1 & 72.4 \\
                OC-SORT~\cite{cao2023observation} & 68.1 & 84.8 & 54.8 & 93.4 & 68.0 \\
                MeMOTR~\cite{gao2023memotr} & 68.8 & 82.0 & 57.8 & 90.2 & 69.9 \\
                MOTIP~\cite{Gao_2025_CVPR} & 72.6 & 83.5 & 63.2 & 92.4 & 77.1 \\
                \rowcolor[gray]{0.95}
                PS-Track (ours) & 45.2 & 50.9 & 41.6 & 30.2 & 46.9 \\
                \midrule
                \multicolumn{6}{l}{\textit{with extra data:}} \\
                GTR~\cite{zhou2022global} & 54.5 & 64.8 & 45.9 & 67.9 & 55.8 \\
                CenterTrack~\cite{zhou2020tracking} & 62.7 & 82.1 & 48.0 & 90.8 & 60.0 \\
                ByteTrack~\cite{zhang2022bytetrack} & 62.8 & 77.1 & 51.2 & 94.1 & 69.8 \\
                TransTrack~\cite{sun2020transtrack} & 68.9 & 82.7 & 57.5 & 92.6 & 71.5 \\
                OC-SORT~\cite{cao2023observation} & 71.9 & 86.4 & 59.8 & 94.5 & 72.2 \\
                DiffMOT~\cite{lv2024diffmot} & 72.1 & 86.0 & 60.5 & 94.5 & 72.8 \\
                \bottomrule
            \end{tabular}
        }
    \end{minipage}
    \vspace{-1em}
    \hfill %
    \begin{minipage}[t]{0.496\linewidth}
        \centering
        \caption{Quantitative comparison on the JRDB test set~\cite{martin2021jrdb}.}
        \vspace{-1em}
        \label{tab:sota_JRDB}
        \footnotesize 
        \setlength{\tabcolsep}{2pt} 
        \resizebox{\linewidth}{!}{
            \begin{tabular}{lcccc} 
                \toprule
                 Method & HOTA$\uparrow$ & OSPA$\downarrow$ & IDF1$\uparrow$ & MOTA$\uparrow$ \\
                \midrule
                 TrackFormer~\cite{meinhardt2022trackformer} & 19.16 & 0.95 & 19.66 & 17.79 \\
                 MOTRv2~\cite{zhang2023motrv2}               & 18.22 & 0.93 & 19.30 & 12.30 \\
                 MeMOTR~\cite{gao2023memotr}                 & 25.10 & 0.87 & 27.46 & 22.53 \\
                 OmniTrack$_{E2E}$~\cite{luo2025omnidirectional} & 21.56 & 0.94 & 22.87 & 25.01 \\
                 OmniTrack++$_{E2E}$~\cite{luo2025omnitrack++}& 25.50 & 0.88 & 28.00 & 21.02 \\
                \midrule
                 SORT~\cite{bewley2016simple}                & 23.49 & 0.90 & 26.11 & 24.59 \\
                 DeepSORT~\cite{wojke2017simple}             & 22.15 & 0.95 & 23.46 & 24.88 \\
                 ByteTrack~\cite{zhang2022bytetrack}         & 25.00 & 0.86 & 27.95 & 26.59 \\
                 Bot-SORT~\cite{aharon2206bot}               & 22.90 & 0.91 & 24.27 & 23.08 \\
                 OC-SORT~\cite{cao2023observation}           & 25.04 & 0.84 & 27.89 & 25.64 \\
                 HybridSORT~\cite{yang2024hybrid}            & 25.01 & 0.85 & 27.82 & 25.03 \\
                 DiffMOT~\cite{lv2024diffmot}                & 19.96 & 0.95 & 20.26 & 20.05 \\
                 OmniTrack$_{DA}$~\cite{luo2025omnidirectional}& 26.92 & 0.84 & 30.26 & 26.60 \\
                 OmniTrack++$_{DA}$~\cite{luo2025omnitrack++}& 27.03 & 0.81 & 29.52 & 25.05 \\
                \rowcolor[gray]{0.95}
                 PS-Track (ours)   & 20.72 & 0.92 & 22.21 & 12.56  \\
                \bottomrule
            \end{tabular}
        }
    \end{minipage}
    \vspace{-1em}
\end{table*}
\subsection{Benchmarking and Comparative Analysis}

\textbf{Performance on Challenging Motion Scenarios.}
We evaluate our method on DanceTrack~\cite{sun2022dancetrack} and SportsMOT~\cite{cui2023sportsmot}, featuring severe non-linear motions and highly uniform appearances. On DanceTrack (Tab.~\ref{tab:dancetrack_val}), PS-Track achieves a remarkable 52.3 HOTA and 53.4 IDF1. Despite strictly utilizing point-level annotations, it surpasses classic fully bounding-box supervised methods, yielding absolute HOTA improvements of +10.5 over CenterTrack~\cite{zhou2020tracking} and +12.6 over FairMOT~\cite{zhang2021fairmot}. On SportsMOT (Tab.~\ref{tab:mot_benchmark}), PS-Track establishes a strong point-supervised baseline with 45.2 HOTA, demonstrating robustness against rapid pose deformations. 
We attribute this efficacy to our PEWA and UGL. PEWA hallucinates structural edges in the frequency domain to counter motion blur, while UGL's dynamic weighting prevents overfitting to occlusion-induced noise.

\noindent\textbf{Performance on Embodied Panoramic Scenarios.} 
We further validate our framework on JRDB~\cite{martin2021jrdb} and EmboTrack~\cite{luo2025omnitrack++}, characterized by first-person ego-motion, severe scale variations, and dense crowds. Acquiring dense bounding boxes here is prohibitively expensive, highlighting our paradigm's practical value. On JRDB (Tab.~\ref{tab:sota_JRDB}), PS-Track achieves 20.72 HOTA, successfully outperforming fully-supervised counterparts like TrackFormer~\cite{meinhardt2022trackformer} (19.16) and DiffMOT~\cite{lv2024diffmot} (19.96). More impressively, on the EmboTrack QuadTrack dataset (Tab.~\ref{tab:sota_compact}), PS-Track delivers a robust 33.9 HOTA and 38.6 IDF1, significantly surpassing fully-supervised models like ByteTrack~\cite{zhang2022bytetrack} (20.7 HOTA) and OC-SORT~\cite{cao2023observation} (20.8 HOTA). This strong performance in ego-centric environments validates our TFP mechanism. By injecting negative spatial cues, TFP constructs a semantic firewall between overlapping pedestrians, resolving the identity merging ambiguity inherent to point-supervised learning in dense crowds.

\begin{table*}[!t]
    \centering
    \caption{Comparison with state-of-the-art methods on EmboTrack~\cite{luo2025omnitrack++}. }
    \label{tab:sota_compact}
    \vspace{-2mm}
    \footnotesize 

    \resizebox{.88\textwidth}{!}{
    \begin{tabularx}{\textwidth}{l YYYY | YYYY}
        \toprule
        \multirow{2}{*}{\textbf{Method}} & \multicolumn{4}{c|}{\textbf{QuadTrack Dataset}} & \multicolumn{4}{c}{\textbf{BipTrack Dataset}} \\
        \cmidrule(lr){2-5} \cmidrule(lr){6-9}
         & \scriptsize HOTA$\uparrow$ & \scriptsize OSPA$\downarrow$ & \scriptsize IDF1$\uparrow$ & \scriptsize MOTA$\uparrow$ & \scriptsize HOTA$\uparrow$ & \scriptsize OSPA$\downarrow$ & \scriptsize IDF1$\uparrow$ & \scriptsize MOTA$\uparrow$ \\
        \midrule
        
        TrackFormer~\cite{meinhardt2022trackformer} & 19.6 & 0.97 & 17.8 & 3.2 & -- & -- & -- & -- \\
        MOTRv2~\cite{zhang2023motrv2}       & 16.4 & 0.96 & 17.1 & -0.1 & 39.3 & 0.78 & 38.6 & 2.7 \\
        MeMOTR~\cite{gao2023memotr}         & --   & --   & --   & --   & 43.2 & 0.82 & 46.2 & 27.9 \\
        OmniTrack++$_{E2E}$~\cite{luo2025omnitrack++} & 34.9 & 0.85 & 41.2 & 18.7 & 44.6 & 0.84 & 46.8 & 21.6 \\

        SORT~\cite{bewley2016simple}        & 14.6 & 0.98 & 15.6 & 4.8  & 42.7 & 0.86 & 45.0 & 28.3 \\
        DeepSORT~\cite{wojke2017simple}     & 21.2 & 0.96 & 22.6 & 5.1  & 41.2 & 0.90 & 38.6 & 22.6 \\
        ByteTrack~\cite{zhang2022bytetrack} & 20.7 & 0.94 & 22.6 & 8.7  & 44.1 & 0.84 & 46.3 & 20.6 \\
        Bot-SORT~\cite{aharon2206bot}       & 15.8 & 0.99 & 15.7 & 5.9  & 42.5 & 0.86 & 40.9 & 25.7 \\
        OC-SORT~\cite{cao2023observation}   & 20.8 & 0.94 & 22.6 & 7.7  & 40.9 & 0.87 & 40.3 & 0.5 \\
        HybridSORT~\cite{yang2024hybrid}    & 16.6 & 0.96 & 17.4 & 6.8  & 42.8 & 0.85 & 43.2 & 13.0 \\
        DiffMOT~\cite{lv2024diffmot}        & 16.4 & 0.97 & 16.6 & 6.2  & 39.3 & 0.95 & 34.3 & 24.5 \\
        OmniTrack++$_{DA}$~\cite{luo2025omnitrack++} & 36.1 & 0.82 & 42.8 & 21.9 & 45.0 & 0.76 & 47.4 & 21.5 \\
        \rowcolor[gray]{0.95}
        PS-Track (ours)          & 33.9 & 0.94 & 38.6 & -1.3 & 33.9 & 0.97 & 32.7 & 14.7 \\
        \bottomrule
    \end{tabularx}
    }
    \vskip -2ex
\end{table*}

\begin{table}[h]
    \centering
    \caption{Ablation study on different associate method using \textbf{Point2RBox-v3}~\cite{zhang2025point2rbox} as the point-supervised detector. All methods are trained with point annotations only.}
    \label{tab:ablation_tbd}
    \vspace{-2mm}
    \footnotesize
    \setlength{\tabcolsep}{4pt}
    \resizebox{.85\textwidth}{!}{
    \begin{tabular}{l|c|ccccc}
        \toprule
        Tracker (TBD Paradigm) & Detection & HOTA$\uparrow$ & DetA$\uparrow$ & AssA$\uparrow$ & MOTA$\uparrow$ & IDF1$\uparrow$ \\
        \midrule
        SORT~\cite{bewley2016simple}        &\multirow{4}{*}{\rotatebox[origin=c]{90}{
    \shortstack[c]{Point2\\RBox-v3} }} & 11.8 & 16.3 & 10.2 & -37.3 & 6.4 \\
        ByteTrack~\cite{zhang2022bytetrack} & & 10.8 & 14.2 & 9.5 & -29.8 & 5.7 \\
        OC-SORT~\cite{cao2023observation}   & & 10.8 & 14.3 & 9.5 & -28.1 & 6.0 \\
        HybridSORT~\cite{yang2024hybrid}    & & 10.8 & 14.3 & 9.5 & -28.1 & 5.9 \\
        \rowcolor[gray]{0.95}
        \textbf{BYTE~\cite{zhang2022bytetrack}+PS-Track (ours)}            & YOLOX~\cite{yolox2021} & 42.9 & 62.6 & 29.5 & 77.3 & 47.9  \\
        \bottomrule
    \end{tabular}
    }
    \vskip -2ex
\end{table}
\subsection{Ablation Studies}
\noindent\textbf{Necessity of the Unified PS-MOT Framework.}
As a baseline, Tab.~\ref{tab:ablation_tbd} evaluates a naive Tracking-by-Detection (TBD) pipeline coupling the point-supervised Point2RBox-v3~\cite{zhang2025point2rbox} with off-the-shelf associators. This decoupled paradigm catastrophically fails (10.8 HOTA, MOTA ${<}-28.0$), as isolated detectors cannot resolve intrinsic scale ambiguity, producing boxes too degraded for any associator to salvage. Conversely, training a YOLOX~\cite{yolox2021} detector via our PS-Track paradigm with BYTE~\cite{zhang2022bytetrack} yields a remarkable 42.9 HOTA and 47.9 IDF1. By unifying perception and temporal evolution, TFP and PEWA explicitly constrain scale generation using motion priors.
This $+32.1$ HOTA margin demonstrates that PS-Track seamlessly integrates into existing TBD frameworks, resolving point-induced scale ambiguity to unlock robust tracking.

\begin{table}[!t]
\centering
\caption{Comprehensive ablation study on the DanceTrack val dataset. 
(a) Contribution of core components. 
(b) Robustness to spatial annotation noise (simulating human clicking variance). 
(c) Sensitivity to the adaptive likelihood weight $\lambda_{reg}$. 
(d) Convergence analysis over training epochs.}
\label{tab:full_ablation}
\vskip -4ex
\footnotesize
\setlength{\tabcolsep}{4pt}

\begin{subtable}{\textwidth}
\centering
\caption{Effectiveness of core components under the full 10-epoch training schedule.}
\vspace{-1em}
\resizebox{.8\textwidth}{!}{
\begin{tabular}{c|ccc|ccccc}
\toprule
Experiments & TFP & PEWA & UGL & HOTA$\uparrow$ & DetA$\uparrow$ & IDF1$\uparrow$ & MOTA$\uparrow$ & AssA $\uparrow$ \\
\midrule
\ding{172} & - & - & - & 30.3 & 45.4 & 32.5 & 36.3 & 20.4 \\
\ding{173} & \checkmark & - & - & 49.0 & 65.2 & 51.8 & 66.9 & 37.1\\
\ding{174} & \checkmark & - & \checkmark & 49.4 & 64.7 & 52.8 & 66.7 & 38.0 \\
\midrule
\rowcolor[gray]{0.95}
\ding{175} & \checkmark & \checkmark & \checkmark & \textbf{50.3} & \textbf{65.3} & \textbf{52.9} & \textbf{66.2} & \textbf{39.1} \\
\bottomrule
\end{tabular}}
\end{subtable}

\begin{subtable}{\textwidth}
    \centering
    \caption{Robustness to annotation noise. Spatial perturbation $\sigma$ (in pixels) is added to the GT points.}
    \vspace{-1em}
    \label{tab:ablation_noise}
    \resizebox{.55\textwidth}{!}{
        \begin{tabular}{l|ccccc}
            \toprule
            Noise Level $\sigma$ & HOTA$\uparrow$ & DetA$\uparrow$ & IDF1$\uparrow$ & MOTA$\uparrow$ & AssA $\uparrow$ \\
            \midrule
            \rowcolor[gray]{0.95}
            $\sigma = 0$ (Clean) & 44.2 & 64.7 & 45.5 & 67.3 & 30.4 \\
            $\sigma = 8$ px      & 44.1 & 65.0 & 45.9 & 68.9 & 30.1 \\
            $\sigma = 16$ px     & 44.1 & 64.5 & 45.1 & 67.4 & 30.4 \\
            $\sigma = 32$ px     & 42.8 & 62.4 & 45.3 & 68.1 & 29.5 \\
            $\sigma = 48$ px     & 39.8 & 60.6 & 41.0 & 64.0 & 26.2 \\
            \bottomrule
        \end{tabular}
    }
\end{subtable}

\begin{subtable}{.8\textwidth}
\centering
\begin{minipage}{.6\textwidth}
\centering
\caption{Loss weight sensitivity.}
\vspace{-1em}
\resizebox{\linewidth}{!}{
\begin{tabular}{c|ccccc}
\toprule
$\lambda_{reg}$ &  HOTA$\uparrow$ & DetA$\uparrow$ & IDF1$\uparrow$ & MOTA$\uparrow$ & AssA $\uparrow$ \\
\midrule
\rowcolor[gray]{0.95}
1  & 44.2 & 64.7 & 45.5 & 67.3 & 30.4 \\
2  & 42.7 & 65.9 & 43.7 & 69.7 & 27.9 \\
3  & 44.8 & 64.9 & 45.5 & 65.9 & 31.1 \\
4  & 40.8 & 64.2 & 41.8 & 65.5 & 26.1 \\
5  & 43.7 & 65.0 & 45.1 & 67.2 & 29.6 \\
\bottomrule
\end{tabular}
}
\end{minipage}
\hfill
\begin{minipage}{.32\textwidth}
\centering
\caption{Epochs analysis.}
\vspace{-1em}
\resizebox{\linewidth}{!}{
\begin{tabular}{c|cc}
\toprule
Epoch & HOTA$\uparrow$ & DetA$\uparrow$ \\
\midrule
2 & 44.2 & 64.7 \\
4 & 46.1 & 63.5 \\
6 & 46.4 & 63.3 \\
8 & 48.6 & 64.8 \\
10 & 50.3 & 65.3 \\
\bottomrule
\end{tabular}
}
\end{minipage}

\end{subtable}
\vspace{-2.5ex}
\end{table}
\noindent\textbf{Effectiveness of Proposed Components.}
Tab.~\ref{tab:full_ablation}(a) details the individual module contributions. The naive point-to-box baseline (\ding{172}) performs poorly (30.3 HOTA) due to the lack of temporal and uncertainty constraints. Integrating Temporal-Feedback Prompting (TFP) (\ding{173}) yields a substantial +18.7 HOTA boost, confirming that motion priors and negative spatial cues effectively alleviate identity merging and fragmentation. Adding Uncertainty-Guided Gaussian Learning (UGL) (\ding{174}) further improves HOTA to 49.4 by adaptively down-weighting residual pseudo-label noise. Finally, the full PS-Track (\ding{175}) equipped with PEWA reaches 50.3 HOTA; PEWA hallucinates structural boundaries in the frequency domain, extracting higher-quality embeddings for association.

\noindent\textbf{Robustness to Annotation Variance.} 
To assess resilience against the inherent spatial variance of human clicking, we inject synthetic Gaussian noise ($\sigma \in \{8, 16, 32, 48\}$ pixels) into training point coordinates. Tab.~\ref{tab:full_ablation}(b) demonstrates PS-Track's remarkable stability, maintaining 44.1 and 42.8 HOTA under 16-pixel and extreme 32-pixel perturbations, respectively. This robustness directly validates the synergy of our framework: TFP and PEWA collaboratively anchor imprecise clicks to true semantic boundaries via motion priors and frequency-domain hallucination, while UGL probabilistically down-weights residual spatial errors, preventing overfitting and ensuring practicality.

\noindent\textbf{Convergence and Sensitivity.} Due to computational constraints, prior ablations employ a rapid 2-epoch schedule. Tab.~\ref{tab:full_ablation}(d) validates this protocol: performance scales smoothly from 44.2 HOTA (epoch 2) to 50.3 (epoch 10) without late-stage degradation, confirming our uncertainty-aware framework effectively resists noise memorization inherent to weak supervision. Furthermore, Tab.~\ref{tab:full_ablation}(c) demonstrates stability across varying regression loss weights ($\lambda_{reg}$), peaking at 44.8 HOTA when $\lambda_{reg}=3$. This proves PS-Track is highly robust and does not rely on exhaustive hyperparameter tuning.

\begin{table}[!t]
    \centering
    \caption{Generalization across diverse tracking paradigms evaluated on the DanceTrack \textit{val} set. \textit{B} and \textit{P} denote fully bbox and point-level supervision, respectively.}
    \vspace{-1em}
    \label{tab:generalization}
    \resizebox{.8\textwidth}{!}{
    \begin{tabular}{l|lc|ccccc}
    \toprule
    Paradigm & Method & Sup. & HOTA$\uparrow$ & DetA$\uparrow$ & IDF1$\uparrow$ & MOTA$\uparrow$ & AssA$\uparrow$ \\
    \midrule

    \multirow{2}{*}{\shortstack[l]{Association-based\\(\textit{e.g.}, SORT~\cite{bewley2016simple})}} 
    & Baseline (BYTE~\cite{zhang2022bytetrack}) & \textit{B} & 47.1 & 70.5 & 51.9 & 88.2 & 31.5  \\
    & \cellcolor[gray]{0.92} BYTE+PS-Track (ours) & \cellcolor[gray]{0.92} \textit{P} & \cellcolor[gray]{0.92} 42.9 & \cellcolor[gray]{0.92} 62.6 & \cellcolor[gray]{0.92} 47.9 & \cellcolor[gray]{0.92} 77.3 & \cellcolor[gray]{0.92} 29.5 \\
    
    \cmidrule(lr){1-8}

    \multirow{2}{*}{\shortstack[l]{Autoregressive\\(\textit{e.g.}, AR-MOT~\cite{jia2026ar})}} 
    & Baseline (AR-MOT~\cite{jia2026ar}) & \textit{B} & 42.4 &70.4 &38.5  &77.6 &25.9  \\
    & \cellcolor[gray]{0.92} AR-MOT+PS-Track (ours) & \cellcolor[gray]{0.92} \textit{P} & \cellcolor[gray]{0.92} 36.9 & \cellcolor[gray]{0.92} 76.4 & \cellcolor[gray]{0.92} 31.0 & \cellcolor[gray]{0.92} 84.5 & \cellcolor[gray]{0.92} 18.0 \\
    
    \cmidrule(lr){1-8}

    \multirow{2}{*}{\shortstack[l]{Query-based\\(\textit{e.g.}, MOTR~\cite{zeng2021motr})}} 
    & Baseline (MOTIP~\cite{Gao_2025_CVPR}) & \textit{B} & 62.2 & 75.3  & 64.8 & 85.2  & 51.5  \\
    & \cellcolor[gray]{0.92} MOTIP+PS-Track (ours) & \cellcolor[gray]{0.92} \textit{P} & \cellcolor[gray]{0.92} 50.3 & \cellcolor[gray]{0.92} 65.3 & \cellcolor[gray]{0.92} 52.9 & \cellcolor[gray]{0.92} 66.2 & \cellcolor[gray]{0.92} 39.1 \\
    \bottomrule
    \end{tabular}}
    \vspace{-1em}
\end{table}

\noindent\textbf{Versatility Across Tracking Paradigms.} 
Tab.~\ref{tab:generalization} demonstrates PS-Track's plug-and-play versatility across three mainstream architectures: Association-based (BYTE~\cite{zhang2022bytetrack}), Autoregressive (AR-MOT~\cite{jia2026ar}), and Query-based (MOTIP~\cite{Gao_2025_CVPR}). 
By integrating our core modules, these fully bounding-box supervised baselines successfully operate under strict point supervision. Notably, BYTE+PS-Track achieves 42.9 HOTA, trailing its fully-supervised counterpart by merely 4.2 points despite drastically reduced annotation costs. Similarly, integrations with MOTIP and AR-MOT maintain effective tracking capabilities, yielding 50.3 and 36.9 HOTA, respectively. These results validate that our framework acts as a universal catalyst—seamlessly mitigating scale ambiguity and unlocking robust point-supervised tracking for diverse pre-existing MOT paradigms.

\subsection{Visualization Analysis}
\begin{figure}[t]
  \centering
  \includegraphics[trim=0mm 0mm 2mm 0mm, clip, width=\textwidth]{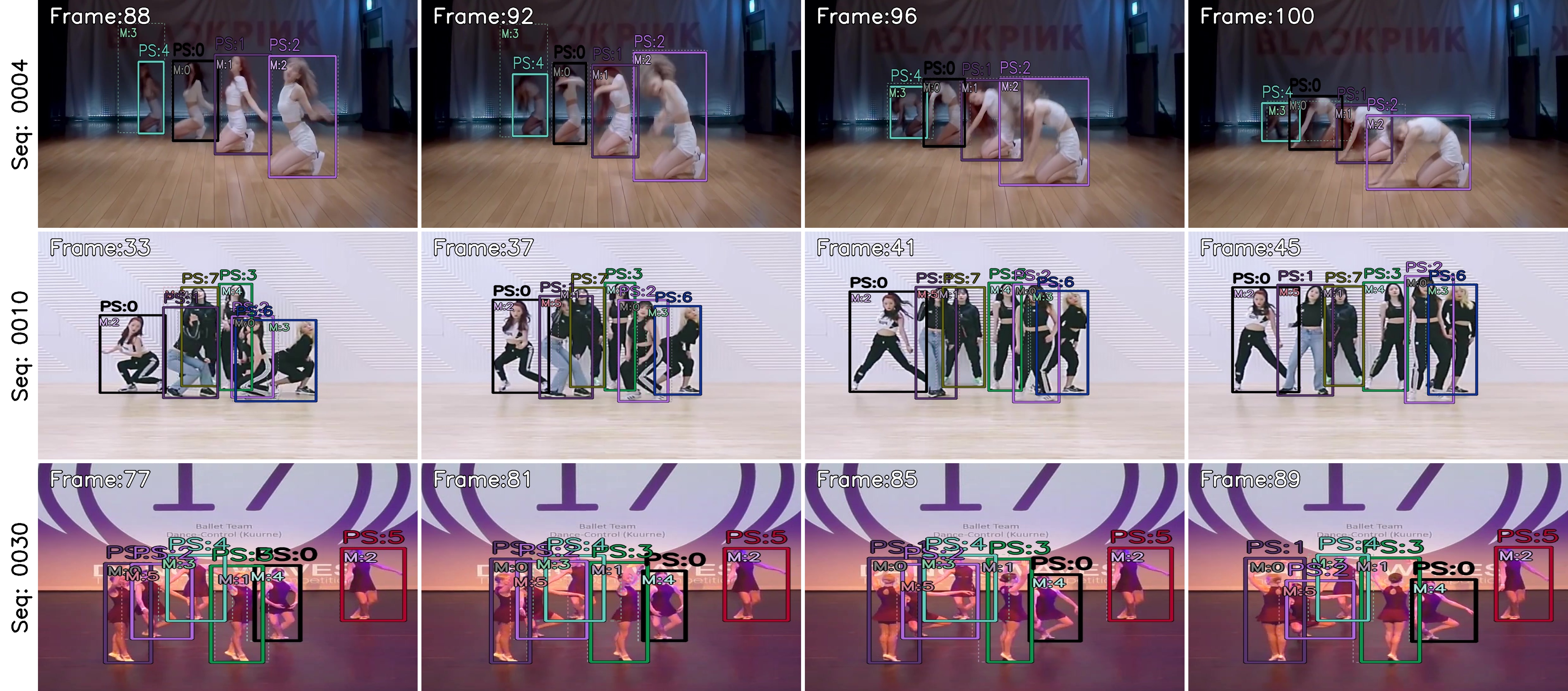}
  \caption{\textbf{Qualitative comparison on the DanceTrack dataset.} Dashed lines with the \textbf{M} prefix denote the tracking results of MOTIP~\cite{Gao_2025_CVPR} while solid lines with the \textbf{PS} prefix represent our proposed PS-Track. 
  }
  \label{fig:visual_comparison_result}
  \vskip -4ex
\end{figure}

Fig.~\ref{fig:visual_comparison_result} qualitatively compares PS-Track with the fully-supervised  MOTIP~\cite{Gao_2025_CVPR} on DanceTrack. Under abnormal poses (seq 003), PS-Track maintains tight localization without bounding box jitter, verifying PEWA's ability to hallucinate structural boundaries in the frequency domain. During dense occlusions (seq 0010), the negative spatial cues in TFP act as a semantic firewall to successfully prevent identity switches caused by mask merging. Furthermore, amid complex intersecting motions of similar targets (seq 0030), our Uncertainty-Guided Gaussian Learning dynamically down-weights noisy visual features to preserve long-term trajectory stability. Overall, despite relying solely on sparse points, PS-Track demonstrates localization and association robustness visually commensurate with fully-supervised paradigms.

\section{Conclusion}

In this paper, we address the scalability bottleneck of Multi-Object Tracking by introducing Point-supervised MOT (PS-MOT). To overcome the inherent spatial ambiguity of sparse points, we propose a unified framework that orchestrates a coarse-to-fine evolution across the data, model, and loss levels. Specifically, the Temporal-Feedback Prompting (TFP) module stabilizes pseudo-labels, the Point-Excited Wavelet Attention (PEWA) module hallucinates sharp object boundaries, and the Uncertainty-Guided Gaussian Learning (UGL) adaptively handles supervision noise. Extensive experiments on four challenging benchmarks, including the panoramic EmboTrack, demonstrate that PS-Track substantially narrows the gap to fully supervised counterparts across diverse mainstream tracking architectures, while reducing measured human annotation time by approximately $64\%$ in our 600-frame setting with offline pseudo-label generation. We hope this work provides a pioneering foundation for future research in label-efficient embodied perception and large-scale tracking.

\section*{Acknowledgments}
This work was supported in part by the National Natural Science Foundation of China (Grant No. 62473139), in part by the Hunan Provincial Research and Development Project (Grant No. 2025QK3019), and in part by the State Key Laboratory of Autonomous Intelligent Unmanned Systems (the opening project number ZZKF2025-2-10). 
The project is partially funded by the Deutsche Forschungsgemeinschaft (DFG, German Research Foundation) – SFB 1574 – 471687386.

\bibliographystyle{splncs04}
\bibliography{main}

\setcounter{section}{0}
\setcounter{figure}{0}
\setcounter{table}{0}
\setcounter{equation}{0}
\renewcommand{\thesection}{\Alph{section}}
\renewcommand{\thefigure}{S\arabic{figure}}
\renewcommand{\thetable}{S\arabic{table}}
\renewcommand{\theequation}{S\arabic{equation}}

\appendix

\clearpage

\section{Implementation Details}
\label{sec:supp_details}
\noindent\textbf{Detailed Network Architecture.} 
We implement PS-Track built upon the MOTIP~\cite{Gao_2025_CVPR} framework, adopting Deformable DETR~\cite{zhu2020deformable} as our core detector. 
The visual features are extracted using a ResNet-50 backbone pre-trained on ImageNet. 
To construct a robust multi-scale representation, we utilize feature maps from the last three backbone stages (which are mapped to $4$ feature levels), and process them using a standard $6$-layer transformer encoder and a $6$-layer transformer decoder. The model maintains $300$ object queries to predict dense instances. During training, PEWA is inserted after the backbone to refine multi-scale features using point-guided heatmaps. During inference, this branch is bypassed, and the tracker follows the original MOTIP inference graph without point inputs. This allows PEWA to refine the multi-scale feature maps and hallucinate structural boundaries before they are fed into the transformer encoder.

\noindent\textbf{Training Setup and Data Augmentation.} 
The entire framework is optimized end-to-end using the AdamW optimizer with a base learning rate of $1 \times 10^{-4}$ and a weight decay of $5 \times 10^{-4}$. To stabilize transformer training, we apply a $1$-epoch learning rate warmup and gradient clipping with a maximal norm of $0.1$. 
Following standard practices in transformer-based detection, the learning rate for the ResNet-50 backbone is scaled down by a factor of $0.1$ ($1 \times 10^{-5}$). During training, we apply standard data augmentation techniques, including random horizontal flipping, random cropping, scale jittering (resizing the shortest side of the input images between $480$ and $800$ pixels, capped at a maximum of $1440$ pixels), and color jittering.

\noindent\textbf{Hardware and Training Schedule.} 
All models are trained on a server equipped with $1$ NVIDIA GeForce RTX 5090 GPU, processing video clips of $30$ frames per batch. 
Due to the substantial computational costs—requiring approximately $8$ hours per epoch on the DanceTrack dataset using a single RTX 5090 GPU—all ablation studies in the main text are consistently evaluated at the 2nd epoch. This rapid prototyping schedule serves as a fair and consistent benchmark for convergence analysis. 
For the final state-of-the-art comparisons, the fully converged PS-Track models are trained for $10$ epochs, with the learning rate decayed by a factor of $0.1$ at the $6$-th and $9$-th epochs.

\noindent\textbf{Inference Details.} 
During the evaluation phase, PS-Track operates in a completely point-free manner. 
We strictly adhere to the default hyperparameter settings of the MOTIP tracker, including a detection confidence threshold of $0.3$, a newborn track initialization threshold of $0.6$, and a miss tolerance of $30$ frames for temporal association, without requiring any specialized, dataset-specific fine-tuning. 
This consistent setup elegantly demonstrates that our framework can seamlessly translate the sparse point guidance learned during training into precise bounding box predictions and robust identity associations during inference.


\begin{figure}[!htbp] 
  \centering
  
  \begin{subfigure}{0.98\textwidth}
    \includegraphics[width=\textwidth]{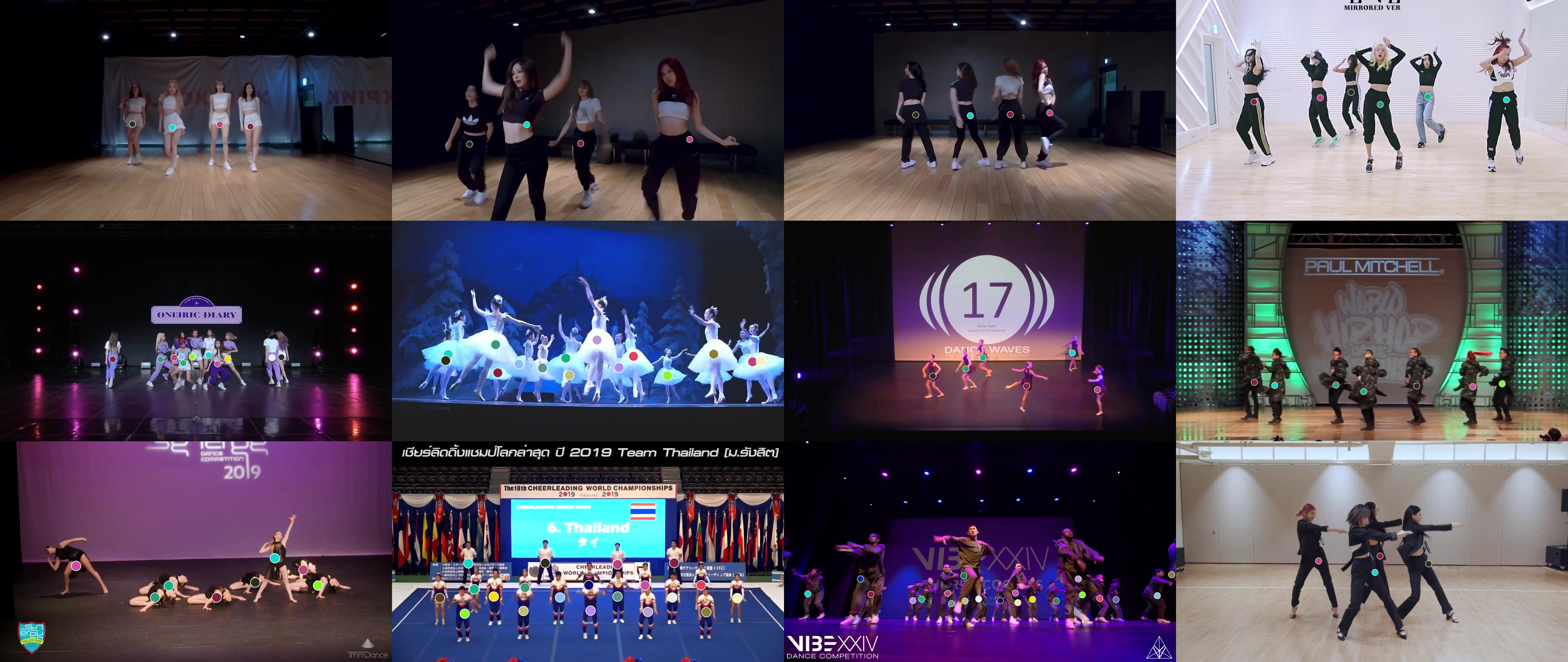}
    \caption{DanceTrack~\cite{sun2022dancetrack}}
    \label{fig:mot_ps_dance}
  \end{subfigure}
  
  \vspace{0.5em} 
  
  \begin{subfigure}{0.98\textwidth}
    \includegraphics[width=\textwidth]{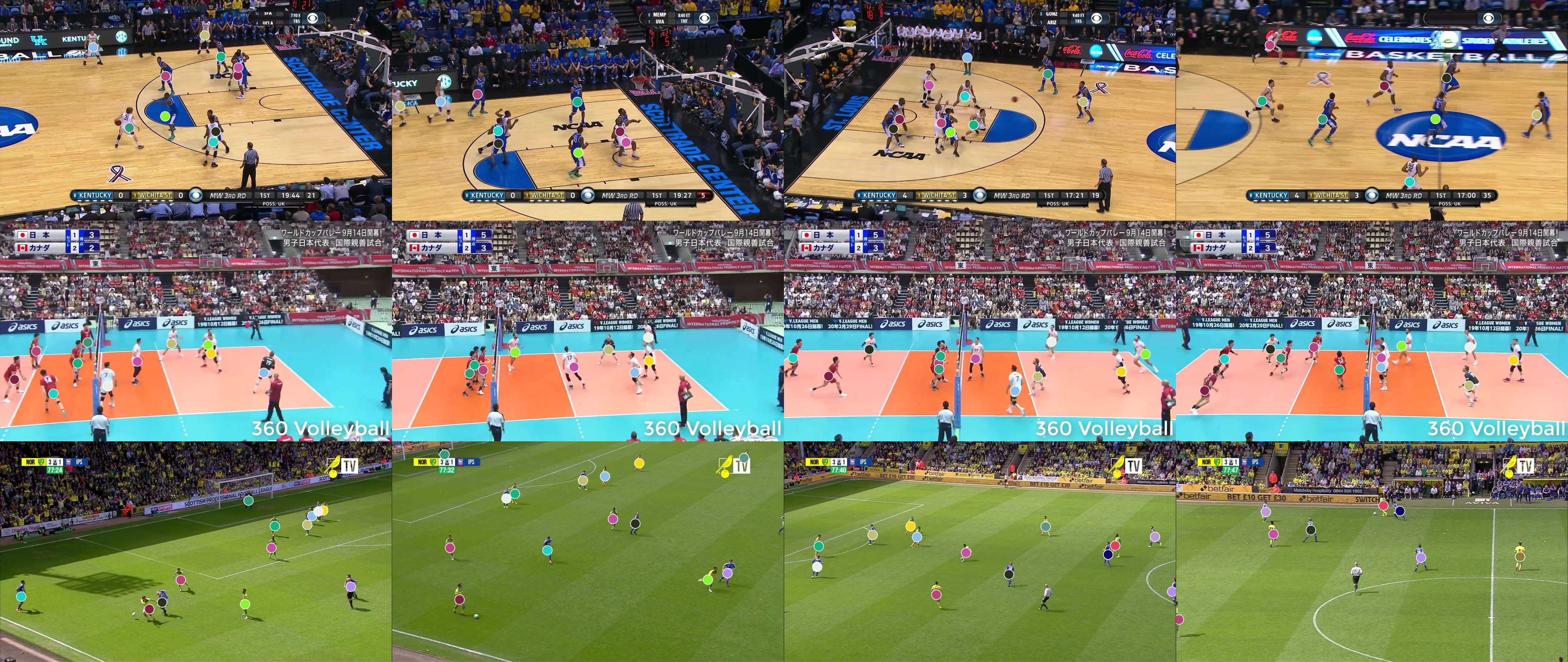}
    \caption{SportsMOT~\cite{cui2023sportsmot}}
    \label{fig:mot_ps_sports}
  \end{subfigure}
  
  \vspace{0.5em}
  
  \begin{subfigure}{0.98\textwidth}
    \includegraphics[width=\textwidth]{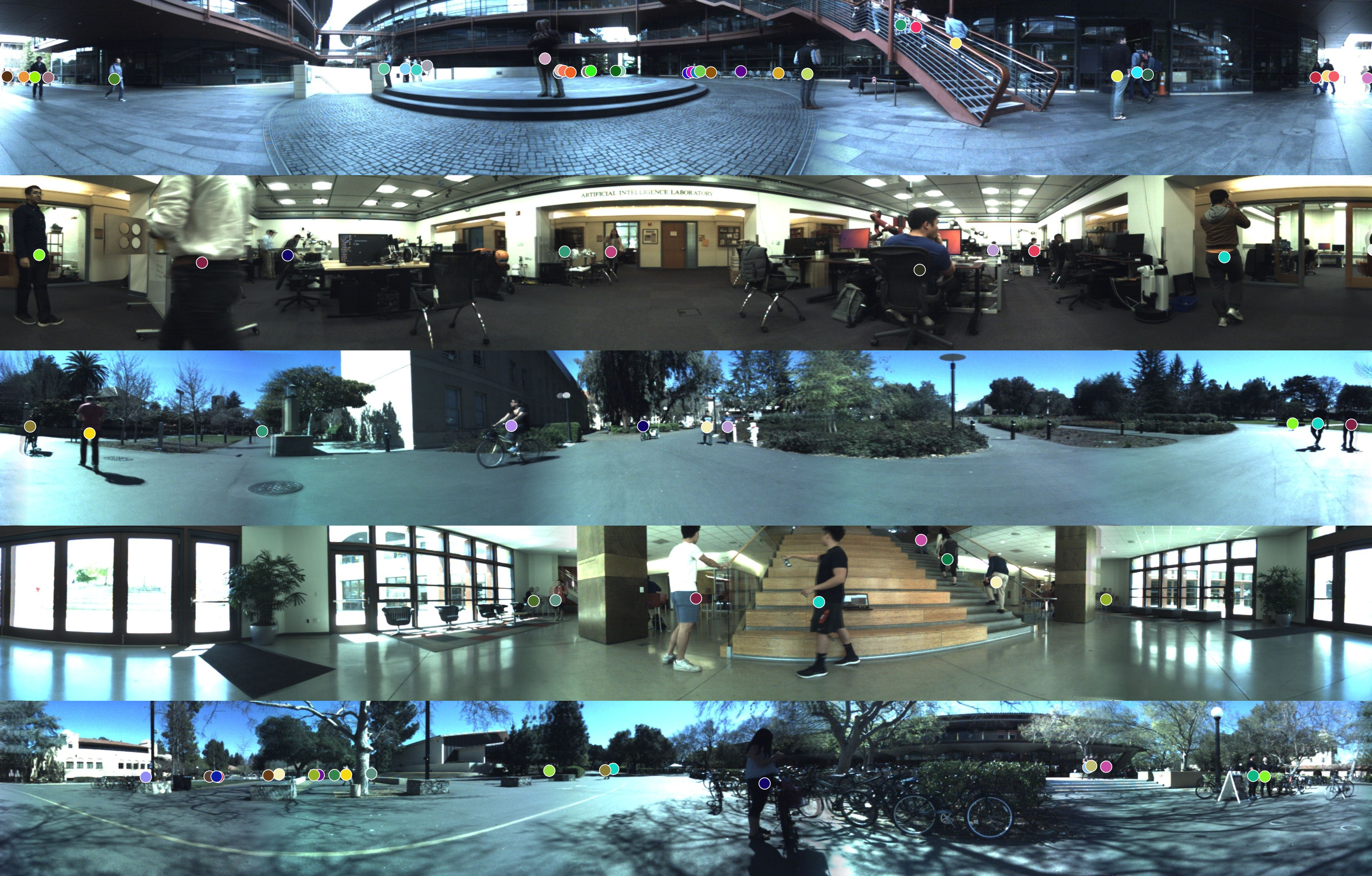}
    \caption{JRDB~\cite{martin2021jrdb}}
    \label{fig:mot_ps_jrdb}
  \end{subfigure}
  
  \caption{\textbf{Visualization of point annotations across diverse datasets.}}
  \label{fig:mot_ps_combined}
\end{figure}

\section{Point Label Generation and Visualization}
\label{sec:supp_point_labels}


\begin{figure}[h]
  \centering
  \includegraphics[trim=0mm 0mm 2mm 0mm, clip, width=\textwidth]{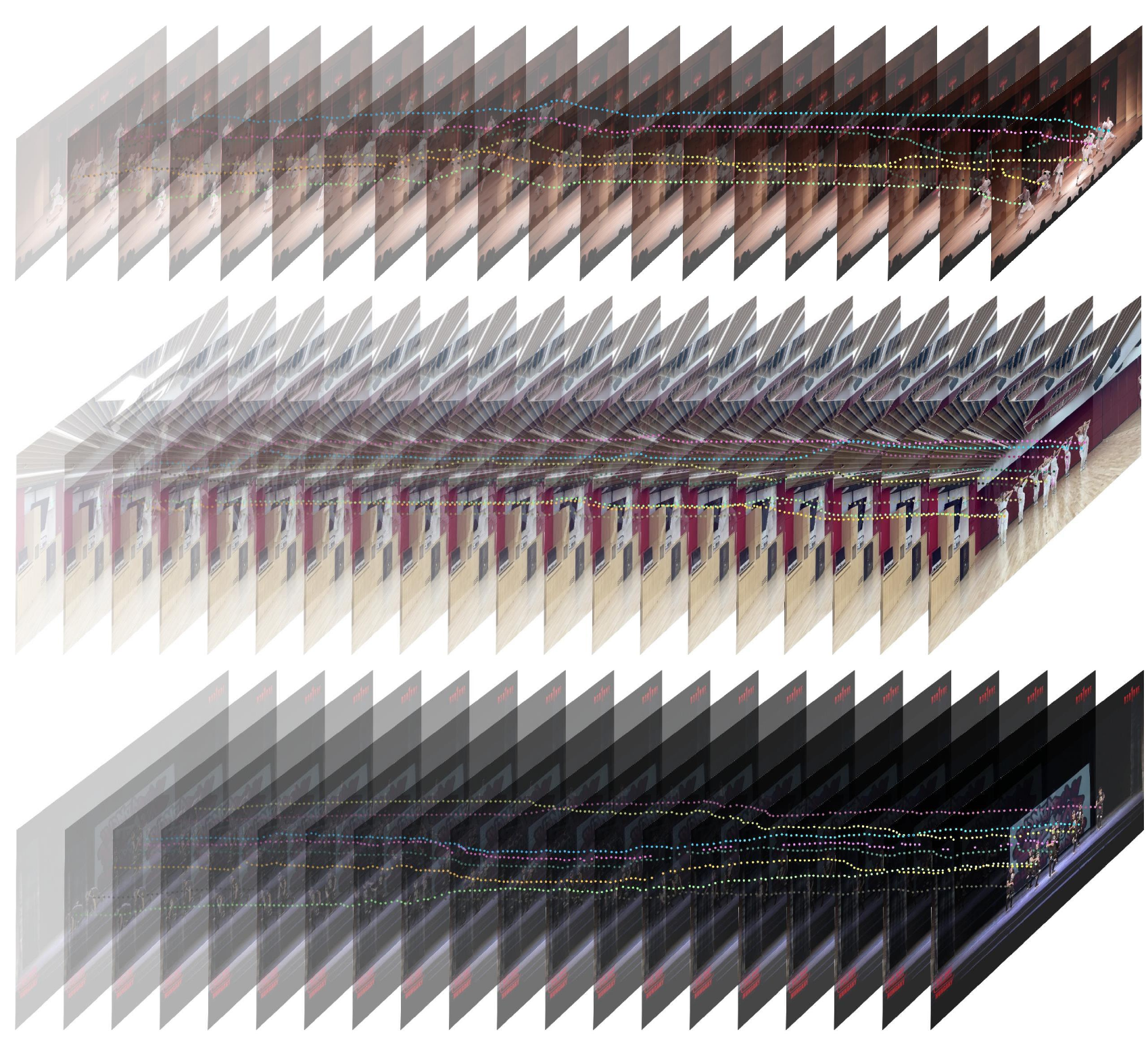}
  \caption{\textbf{Spatio-temporal visualization of point-level training labels on the DanceTrack dataset.} 
  Consecutive frames are sequentially stacked to illustrate the temporal evolution of the annotations.}
  \label{fig:anno}
\end{figure}

To establish the point-level annotations used in our experiments, we synthesize point labels from the centers of the original ground-truth bounding boxes under a controlled benchmarking protocol. This strategy standardizes the supervision signal across datasets and enables reproducible comparisons, while serving as a proxy for point supervision rather than a direct human-click annotation study. Crucially, this generation process is completely automatic and strictly devoid of any subsequent manual curation, heuristic refinement, or human-in-the-loop fine-tuning. Consequently, although center sampling provides a simple and consistent point source, it can still introduce spatial ambiguity in challenging cases. For instance, in scenarios involving severe occlusions, extreme aspect ratios, or complex non-rigid deformations, the geometric center of a bounding box may fall outside the visible physical boundary of the target instance, or inadvertently align with a neighboring object within densely packed or cluttered environments.

We intentionally preserve these noisy artifacts to faithfully simulate a realistic, highly scalable annotation paradigm where human annotators might casually click near a target rather than precisely on it. By subjecting PS-Track to these unpolished, raw point seeds, we rigorously evaluate the framework's capability to cultivate accurate instance awareness and maintain identity consistency under strictly sparse and highly imperfect supervision. 
As visualized in Fig.~\ref{fig:mot_ps_combined} and Fig.~\ref{fig:anno}, our method demonstrates remarkable resilience, effectively interpreting these diverse and occasionally inaccurate point prompts across vastly different scenarios, from tracking dense dancers to dynamic sports players and crowded embodied perspectives under highly challenging and unconstrained conditions.

\section{Addressing the Training-Inference Discrepancy of PEWA}
\label{sec:pewa_discrepancy}

\noindent\textbf{Clarification on the Inference Phase.} 
A natural question regarding the Point-Excited Wavelet Attention (PEWA) module is its behavior during the evaluation phase and the potential risk of label leakage. We explicitly clarify that PEWA is designed strictly as a \textit{training-only} structural prior generator. It utilizes the ground-truth point seeds solely during training to hallucinate high-frequency boundaries, facilitating the learning of optimal representations. During inference, the model operates in a completely point-free manner: it takes only raw images as input, and the entire PEWA module is structurally bypassed and deactivated. Consequently, there is absolutely no reliance on point annotations during evaluation, completely eliminating any possibility of label leakage.

\noindent\textbf{Investigating the Structural Gap.} 
Given that PEWA modulates the feature maps fed into the transformer encoder during training but is absent during inference, this structural discrepancy raises a critical concern: does the network develop a biased over-reliance on these point-guided priors, thereby suffering a catastrophic performance drop when they are removed during inference? To investigate this, we intuitively hypothesized that stochastically activating PEWA with a probability $p$ during training—acting as a structural dropout—might bridge this domain gap. To validate this, we conducted a comprehensive ablation study on the stochastic activation probability $p$ evaluated on the highly dynamic DanceTrack validation set~\cite{sun2022dancetrack}, as reported in Tab.~\ref{tab:ablation_pewa_prob}. Here, $p=0.0$ denotes the baseline without PEWA, while $p=1.0$ indicates consistent activation.
\begin{table}[h]
    \centering
    \caption{Ablation of PEWA activation probability $p$. $p=0.0$ denotes the baseline without PEWA, while $p=1.0$ indicates consistent activation.}
    \label{tab:ablation_pewa_prob}
    \resizebox{.9\textwidth}{!}{
        \begin{tabular}{c|ccccc}
            \toprule
            Activation Prob. $p$ & HOTA$\uparrow$ & DetA$\uparrow$ & IDF1$\uparrow$ & MOTA$\uparrow$ & AssA$\uparrow$ \\
            \midrule
            \rowcolor[gray]{0.95}
            $p = 1.0$ (Ours) & 44.2 & 64.7 & 45.5 & 67.3 & 30.4 \\
            $p = 0.8$ & 43.1 & 64.5 & 44.9 & 68.9 & 28.9 \\
            $p = 0.6$ & 43.6 & 64.0 & 44.8 & 68.5 & 29.8 \\
            $p = 0.4$ & 43.3 & 65.2 & 45.0 & 68.9 & 29.0 \\
            $p = 0.2$ & 43.0 & 64.7 & 44.2 & 66.9 & 28.8 \\
            $p = 0.0$ (Baseline)& 43.7 & 65.0 & 45.1 & 67.2 & 29.6 \\
            \bottomrule
        \end{tabular}
    }
\end{table}

\noindent\textbf{Analysis on Feature Consistency.} 
Contrary to the initial hypothesis, the results demonstrate that introducing stochasticity ($p \in [0.2, 0.8]$) does not bridge the gap but rather degrades the overall tracking performance, particularly the association accuracy (AssA). For instance, compared to the baseline ($p=0.0$), setting $p=0.4$ drops the AssA from $29.6$ to $29.0$. 
We attribute this phenomenon to \textit{feature representation inconsistency}. Since PEWA explicitly reconstructs structural edges in the frequency domain, randomly activating and deactivating it across training iterations causes severe feature jittering. This oscillation confuses the downstream transformer encoder, hindering its ability to learn stable and discriminative temporal identity embeddings.
Conversely, consistent activation ($p=1.0$) yields the optimal performance ($44.2$ HOTA and $30.4$ AssA). 
This confirms that a stable, deterministic structural prior during training is far more beneficial for identity preservation than attempting to explicitly mimic the point-free inference state. 
The network learns a robust, generalized representation from the continuously refined boundaries, naturally translating this capability to the inference phase without suffering from the hypothesized structural bias.

\section{Implementation Details of Comparative Methods}
\label{sec:supp_baselines}

To ensure a fair and rigorous evaluation, we provide comprehensive implementation details for the comparative baselines used in our experiments, specifically addressing the adaptations made for point-supervised detection and the reproduction of unreleased tracking frameworks.

\noindent\textbf{Adaptation of Point2RBox-v3.} 
To construct a state-of-the-art Tracking-by-Detection (TBD) baseline under point-level supervision, we adopt Point2RBox-v3~\cite{zhang2025point2rbox}, which represents the latest advancement in point-supervised object detection to the best of our knowledge. Since the DanceTrack dataset relies on horizontal bounding boxes (HBox), whereas Point2RBox-v3 natively predicts rotated bounding boxes (RBox) parameterized as $[x, y, w, h, \theta]$, we perform a straightforward post-processing alignment. We strictly retain the original network architecture and default parameters without any manual tuning. The only modification made is adapting the data loader to support the DanceTrack dataset format. During the final inference stage, we project the predicted rotated boxes back to horizontal boxes by explicitly forcing the rotation angle $\theta$ to $0^{\circ}$. This minimal intervention ensures seamless compatibility with off-the-shelf temporal associators and standard MOT evaluation protocols while strictly preserving the original detection capability of the baseline.

\noindent\textbf{Reproduction of AR-MOT.} 
As shown in Tab.~7 of the main text, we evaluate the generalization of our PS-Track framework across different tracking paradigms, including the autoregressive architecture represented by AR-MOT~\cite{jia2026ar}. Since the official codebase for AR-MOT is currently unreleased, we carefully reproduced the algorithm based on the original manuscript to conduct this experiment. 
Our implementation faithfully reconstructs its core components, including the Region-Aware Alignment (RAA) and Temporal Memory Fusion (TMF) modules. For certain micro-level hyperparameters not completely detailed in the paper, we assigned values based on standard empirical practices in transformer-based tracking. Crucially, our manual reproduction successfully achieves the tracking accuracy reported in the original AR-MOT paper on the DanceTrack dataset. Therefore, we confidently utilize this validated reproduction to perform the AR-MOT + PS-Track experiments presented in Tab.~7.

\section{Computational Overhead Analysis}

\begin{table}[h]
\centering
\caption{Computational overhead compared with the MOTIP baseline. PEWA and SAM are not used during inference.}
\begin{tabular}{l|ccccc}
\toprule
Method & Params & FLOPs & Train Mem. & Train Time & Inference FPS \\
\midrule
MOTIP & 59.1M & 408.9G & 23.7GB & $\sim$7.4h/ep & 29.82 \\
PS-Track & 61.9M & 408.9G & 24.2GB & $\sim$8.0h/ep & 29.48 \\
\bottomrule
\end{tabular}
\label{tab:s2}
\end{table}

To quantify the additional computational cost introduced by PS-Track, we compare it with the MOTIP baseline under the same training and inference setting. As shown in Tab.~\ref{tab:s2}, PS-Track introduces only moderate training overhead: the training memory increases from 23.7GB to 24.2GB, and the training time increases from approximately 7.4h to 8.0h per epoch. The parameter count rises from 59.1M to 61.9M because it includes the training-time PEWA branch. During inference, however, PS-Track follows the standard MOT pipeline: SAM is used only for offline pseudo-label generation, and PEWA is bypassed. Therefore, the inference FLOPs remain unchanged and the FPS is identical to MOTIP.

\subsection{Effect of Promptable Segmentation Backbones}

To examine the dependence of PS-Track on the quality of offline pseudo-label generation, we replace only the promptable segmentation backbone used in TFP while keeping the tracker architecture, training schedule, and inference protocol unchanged. As shown in Tab.~\ref{tab:sam_variant}, stronger SAM variants produce more reliable pseudo-labels and lead to consistently better tracking performance. This verifies that PS-Track benefits from higher-quality foundation-model priors during training. Meanwhile, all SAM variants are used only for offline pseudo-label generation and are not involved during inference, so the standard point-free MOT inference setting remains unchanged.

\begin{table}[t]
\centering
\caption{Effect of the promptable segmentation backbone used for offline pseudo-label generation. Only the SAM variant in TFP is changed, while the tracker and inference protocol remain unchanged.}
\label{tab:sam_variant}
\begin{tabular}{l|ccccc}
\toprule
SAM Variant & HOTA$\uparrow$ & DetA$\uparrow$ & IDF1$\uparrow$ & MOTA$\uparrow$ & AssA$\uparrow$ \\
\midrule
SAM-v1 & 20.7 & 29.9 & 16.5 & -27.5 & 14.4 \\
SAM-v2 & 39.0 & 53.0 & 42.0 & 44.1 & 29.0 \\
\rowcolor[gray]{0.95}
SAM-v3 (Ours) & 50.3 & 65.3 & 52.9 & 66.2 & 39.1 \\
\bottomrule
\end{tabular}
\end{table}

\section{More Qualitative Results}
\label{sec:supp_vis}

\begin{figure}[h]
  \centering
  \includegraphics[trim=0mm 0mm 2mm 0mm, clip, width=\textwidth]{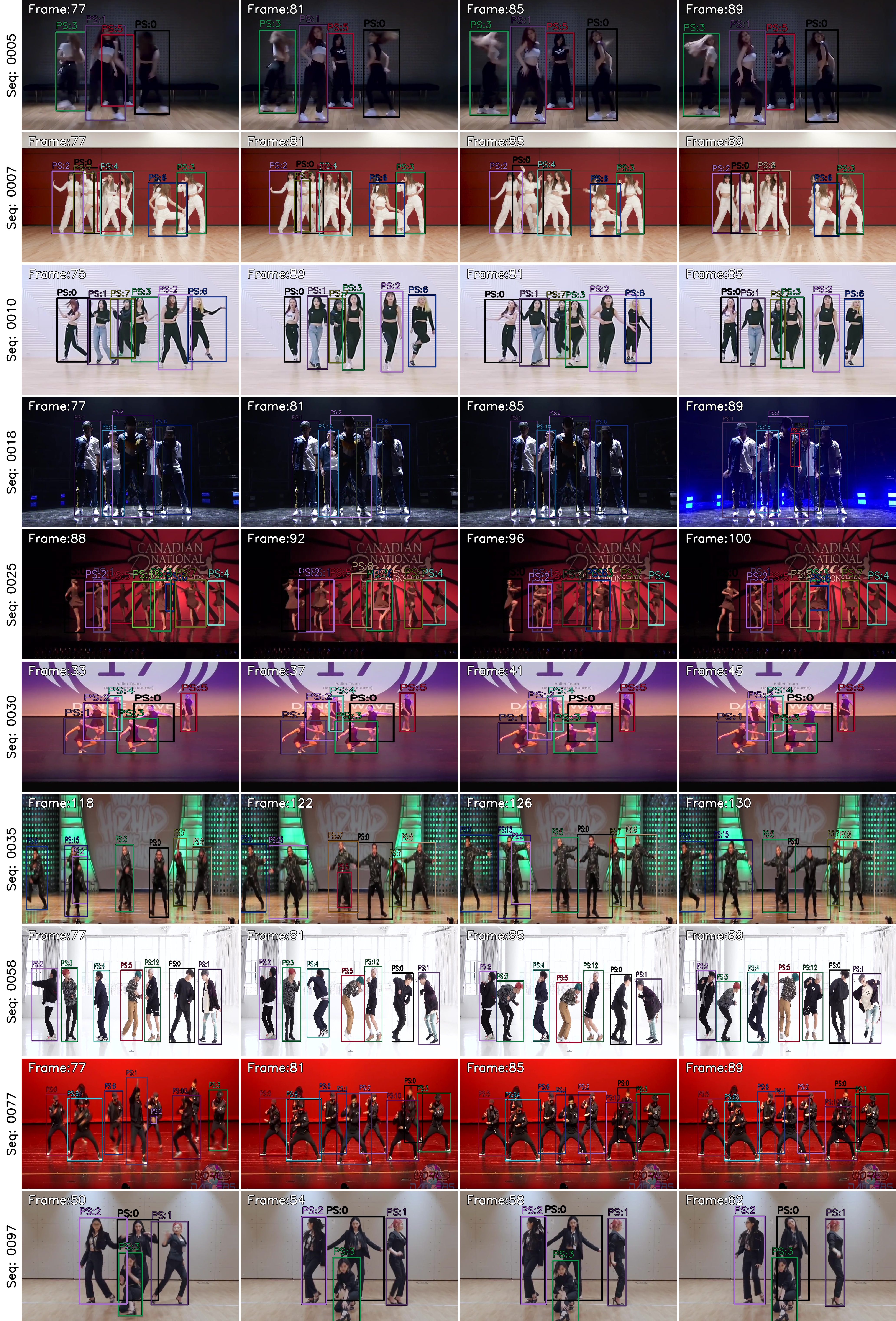}
    \caption{\textbf{Qualitative results on the DanceTrack}~\cite{sun2022dancetrack}\textbf{.} Bounding boxes with the `PS' prefix denote the continuous trajectories predicted by our PS-Track, demonstrating robust localization and stable identity association under purely point-level supervision.}
  \label{fig:dancetrack_track}
\end{figure}

\begin{figure}[h]
  \centering
  \includegraphics[trim=0mm 0mm 2mm 0mm, clip, width=\textwidth]{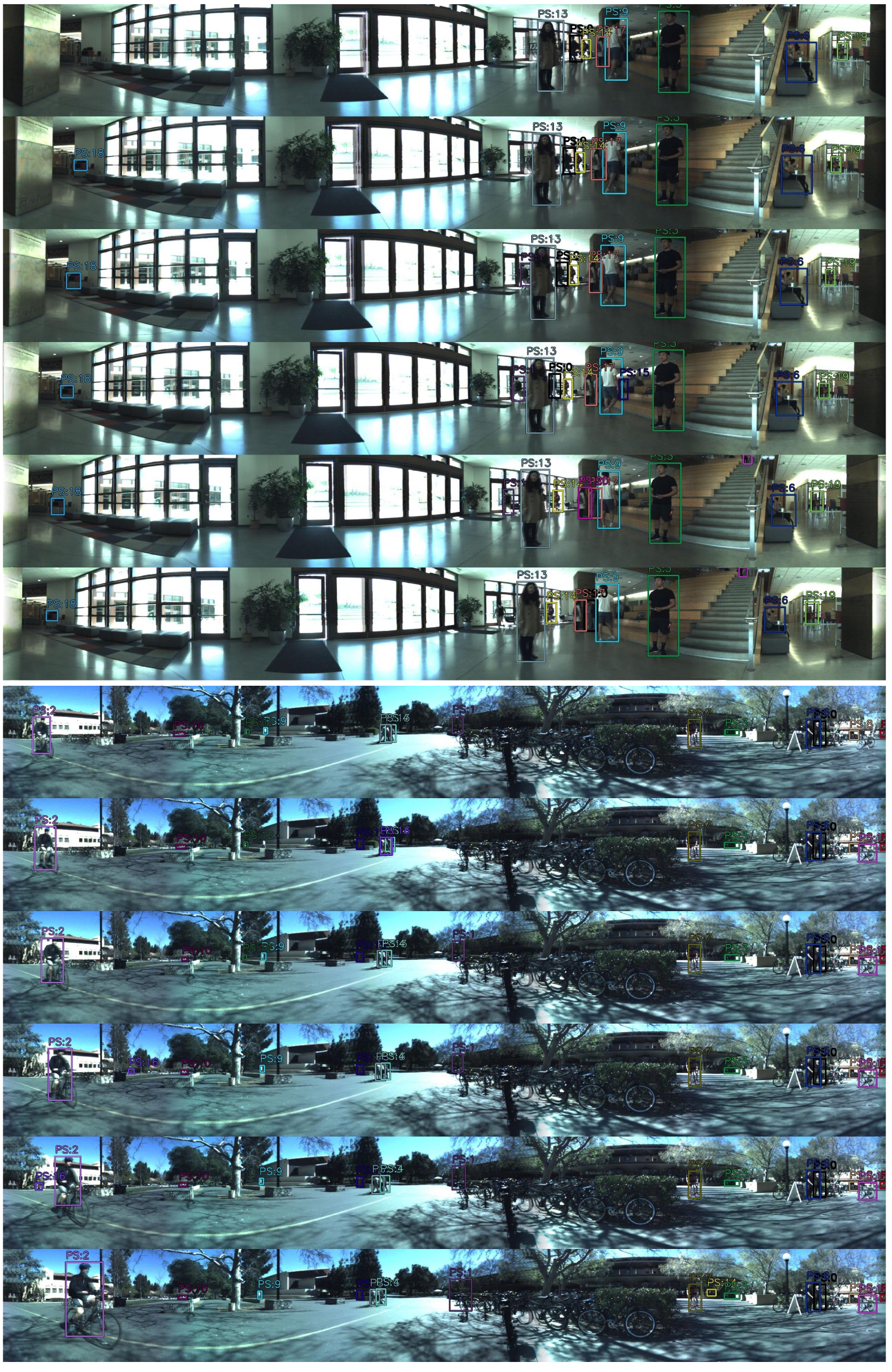}
    \caption{\textbf{Qualitative results on the JRDB dataset}~\cite{martin2021jrdb}\textbf{.} 
    Bounding boxes with the `PS' prefix denote the continuous trajectories predicted by our PS-Track, demonstrating robust localization and stable identity association under purely point-level supervision.}
  \label{fig:jrdb_track}
\end{figure}

\begin{figure}[h]
  \centering
  \includegraphics[trim=0mm 0mm 2mm 0mm, clip, width=\textwidth]{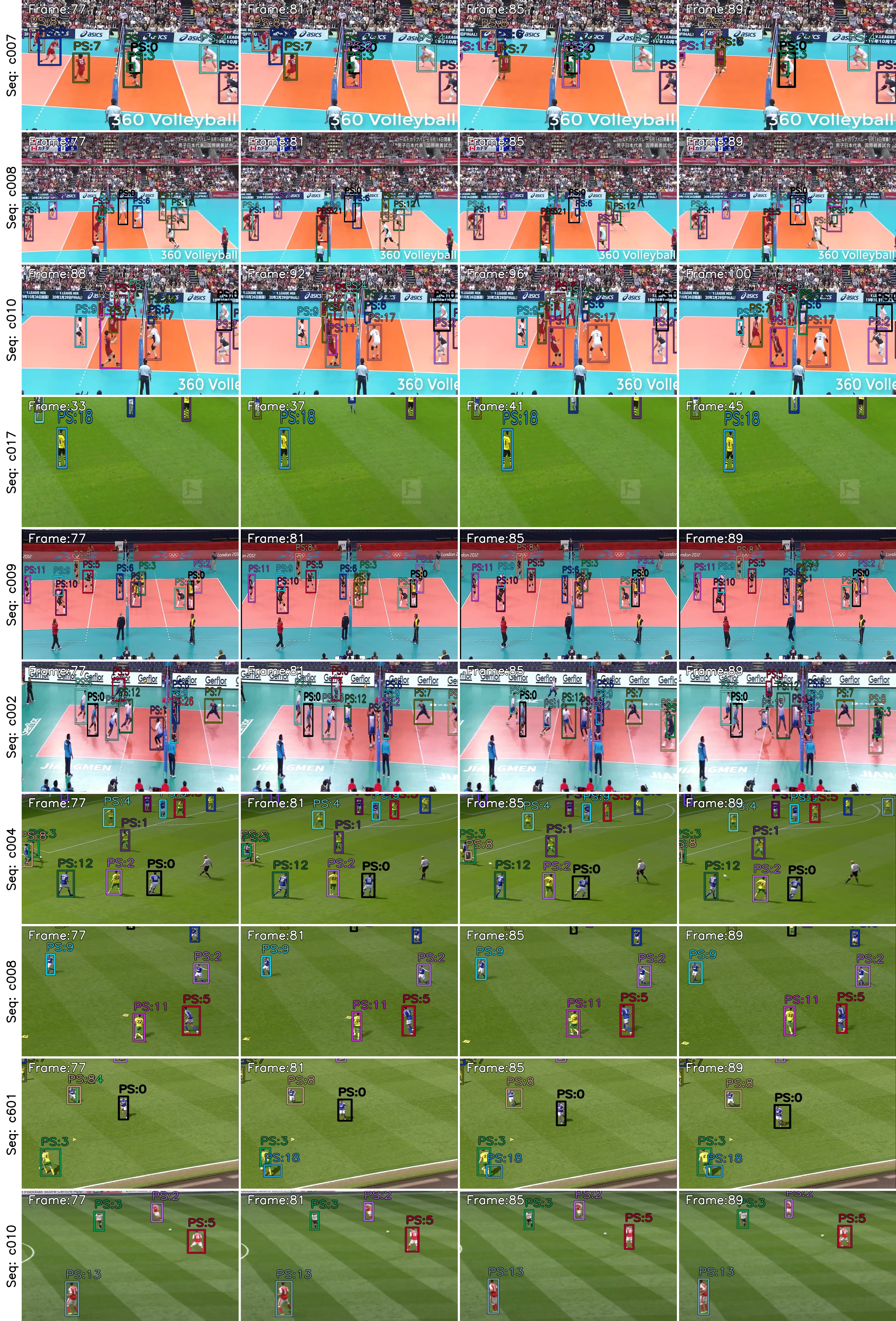}
    \caption{\textbf{Qualitative results on the SportsMOT}~\cite{cui2023sportsmot}\textbf{.} 
    Bounding boxes with the `PS' prefix denote the continuous trajectories predicted by our PS-Track, demonstrating robust localization and stable identity association under purely point-level supervision.}
  \label{fig:sport_track}
\end{figure}

To further intuitively demonstrate the effectiveness and robustness of our PS-Track framework, we provide comprehensive qualitative tracking results across three distinct and challenging datasets: DanceTrack~\cite{sun2022dancetrack}, SportsMOT~\cite{cui2023sportsmot}, and JRDB~\cite{martin2021jrdb} datasets. 
As illustrated in Fig.~\ref{fig:dancetrack_track}, Fig.~\ref{fig:jrdb_track}, and Fig.~\ref{fig:sport_track}, PS-Track maintains highly accurate bounding box localization and identity-consistent tracking trajectories under extremely complex scenarios, despite relying exclusively on sparse point supervision during training. 

Specifically, in DanceTrack scenes characterized by uniform appearances and frequent cross-overs, our model successfully predicts precise bounding boxes to effectively prevent identity switches. Similarly, against the backdrop of rapid camera movements and extreme human pose deformations in SportsMOT, PS-Track exhibits remarkable temporal stability, proving that the point-guided features are highly discriminative. Furthermore, in densely crowded, egocentric scenarios with severe perspective scaling like JRDB, our framework effectively separates closely interacting pedestrians, demonstrating its immense potential and scalability in embodied vision tasks. Collectively, these visualizations confirm that our point-supervised paradigm effectively cultivates robust instance awareness, successfully bridging the gap between sparse annotations and dense perception requirements in dynamic, real-world environments.

\section{Discussion and Limitations}
\label{sec:discussion}
\noindent\textbf{Broader Impacts.} 
The dominant reliance on dense bounding box annotations has long been a scalability bottleneck in the MOT community. By formalizing the Point-Supervised MOT (PS-MOT) paradigm, the impact of our framework extends beyond reducing manual spatial box-fitting effort suggested by prior per-instance annotation studies. Traditional rigid, axis-aligned bounding boxes intrinsically struggle to depict true geometric structures under severe perspective distortions, such as those captured by omnidirectional cameras in embodied navigation or fisheye lenses. In contrast, point-level signals offer stable topological anchors that naturally resist these geometric deformations. 
Consequently, PS-MOT not only facilitates large-scale data curation but also unlocks scalable tracking in geometrically complex scenarios where precise bounding box definition is fundamentally ambiguous or ill-defined. Despite this paradigm shift, cultivating precise dense instance representations exclusively from dimensionless point seeds remains inherently challenging across diverse and unstructured robotic environments.

\noindent\textbf{Limitations and Future Work.} 
Our framework naturally exhibits certain limitations under extreme physical conditions. First, while our Temporal-Feedback Prompting (TFP) successfully employs negative spatial cues to separate adjacent targets, it may still struggle in scenarios involving severe mutual occlusion and highly entangled limbs (\textit{e.g.}, crowded wrestling or dense dance formations). In such cases, a single point prompt might fall onto an ambiguous intersecting boundary, leading the foundation model to generate over-segmented or fragmented pseudo-labels. Second, our Point-Excited Wavelet Attention (PEWA) module explicitly leverages high-frequency components to predict structural boundaries. When objects undergo extreme motion blur, these high-frequency edge features are physically degraded, making it difficult to accurately infer full physical extents without explicit scale priors. Another limitation lies in the annotation protocol. Our benchmark experiments synthesize point labels from GT box centers for controlled and reproducible evaluation. This protocol does not measure real human clicking time and may provide cleaner center cues than practical annotations. Moreover, MOT still requires temporal identity linkage, which is not removed by point supervision. Finally, TFP introduces additional offline pseudo-label generation cost through SAM. Therefore, our annotation-efficiency analysis should be interpreted as a controlled estimate of reduced box-labeling effort rather than a complete end-to-end cost measurement.

To address these challenges, future work could explore integrating multi-modal prompts (\textit{e.g.}, combining sparse points with coarse language descriptions) to resolve semantic ambiguities. Furthermore, while PS-Track drastically reduces spatial annotation costs, it currently relies on frame-by-frame point supervision. A highly promising future direction is to extend this paradigm to temporally sparse annotations—such as providing point prompts only at fixed intervals (\textit{e.g.}, every $5$ frames)—coupled with unsupervised motion propagation, to push the boundaries of label-efficient video perception even further. Beyond 2D domains, adapting this point-driven framework to 3D MOT represents another compelling frontier. Given that annotating dense 3D bounding boxes in LiDAR or multi-view spaces is notoriously labor-intensive, elevating our topological center-driven approach to 3D point clouds or BEV representations could unlock unprecedented scalability for autonomous driving and spatial computing. We hope PS-Track establishes a solid foundation that inspires continued exploration in this domain.

\end{document}